\documentclass{article}

\usepackage{preprint}

\usepackage[utf8]{inputenc}
\usepackage[T1]{fontenc}

\usepackage{hyperref}
\hypersetup{
  pdftitle={Can We Trust In-Distribution Success? Locked Evaluation Reveals Transfer Failure and Sampling-Depth Entanglement in CRISPRi Perturbation Prediction},
  pdfauthor={Mehrdad Shoeibi, Niloofar Yousefi},
  pdfsubject={Machine learning evaluation; single-cell perturbation prediction},
  colorlinks=true,
  linkcolor=black,
  citecolor=black,
  urlcolor=blue
}
\usepackage{url}
\usepackage{booktabs}
\usepackage{amsfonts}
\usepackage{amsmath}
\usepackage{amssymb}
\usepackage{amsthm}
\usepackage{nicefrac}
\usepackage{microtype}
\usepackage{xcolor}
\usepackage{graphicx}
\usepackage{subcaption}
\usepackage{enumitem}
\usepackage{alphalph}
\newcommand{\Rb}{\mathbb{R}}
\newcommand{\Eb}{\mathbb{E}}

\newcommand{\indep}{\perp\!\!\!\perp}
\newcommand{\magmap}{\boldsymbol{m}}
\newcommand{\magmaptilde}{\widetilde{\boldsymbol{m}}}
\newcommand{\xvec}{\mathbf{x}}

\newcommand{\zvec}{\mathbf{z}}
\newcommand{\wvec}{\mathbf{w}}
\newcommand{\spsm}{\textsc{spsm}}
\newcommand{\magspsm}{\textsc{mag-spsm}}
\newcommand{\magrank}{\textsc{mag-Rank}}
\newcommand{\maglinear}{\textsc{mag-Linear}}
\newcommand{\rsq}{R^{2}}
\newcommand{\nmatch}{n_{\mathrm{match}}}
\newcommand{\sigmoid}{\operatorname{sigmoid}}

\title{Can We Trust In-Distribution Success? Locked Evaluation Reveals Transfer Failure and Sampling-Depth Entanglement in CRISPRi Perturbation Prediction}

\author{%
  Mehrdad Shoeibi \\
  Department of Industrial Engineering and Management Systems \\
  University of Central Florida \\
  Orlando, FL, USA \\
  \texttt{Mehrdad.Shoeibi@UCF.edu} \\
  \AND
  Niloofar Yousefi \\
  Department of Industrial Engineering and Management Systems \\
  University of Central Florida \\
  Orlando, FL, USA \\
  \texttt{Niloofar.Yousefi@ucf.edu} \\
}

\begin{document}

\maketitle

\begin{abstract}
AI evaluation can support the wrong inference when an in-domain benchmark success
does not survive distribution shift, or when the benchmark endpoint is entangled
with a design factor. We study this problem in CRISPRi perturbation-effect prediction,
evaluating a frozen Geneformer representation under a locked, pre-registered
protocol---heads and model selection were frozen before test evaluation; the protocol
required external outcome labels to remain withheld until final unblinding; and
analysis-governing decisions were fixed before the evaluations they govern. In-distribution on the Virtual Cell Challenge (VCC),
the frozen representation carries measurable predictive information beyond a
dimension-matched random-feature control ($\Delta\rsq = +0.1645$, $95\%$ CI
$[+0.1375, +0.1920]$), satisfying the pre-registered informativeness gate required
before interpreting transfer. It then fails zero-shot transfer on both external
screens (Spearman $\rho = -0.139$ and $-0.267$), lying below that control on each.
Adding a predefined magnitude block improves the representation externally
($\Delta\rho = +0.032$ and $+0.143$) but, under the frozen primary head, does not
rescue transfer: both remain negative. A pre-registered, count-adjusted max-response
secondary is positively associated with the outcome on both screens; we report it
as correlational and secondary, not as a recovered magnitude signal. Finally, the
VCC endpoint is strongly sample-size associated---a count-only linear model reaches
$\rsq = +0.4325$, versus $+0.2589$ for the four magnitude scalars; adding those
scalars to cell count improves $\rsq$ by only $+0.0017$---so much of the
aggregate-magnitude signal overlaps with cell count. This case study shows how
locking the evaluation, harmonizing the measured endpoint, and separating primary
from secondary evidence can change the inference supported by an AI benchmark.
\end{abstract}

\section{Introduction}
\label{sec:intro}

A trustworthy AI evaluation must justify the inference drawn from a benchmark:
what construct the endpoint measures, whether success survives a relevant distribution
shift, and whether the apparent signal is entangled with properties of the evaluation
design. CRISPRi perturbation-effect prediction provides a concrete scientific case.
Predicting how a cell's transcriptome responds to a genetic perturbation is central to
computational single-cell biology, with applications to target discovery and synthetic
biology \citep{dixit2016perturbseq, replogle2022mapping, lotfollahi2019scgen,
roohani2024gears}. The practically demanding variant emphasized by benchmarks such
as the Virtual Cell Challenge (VCC) \citep{roohani2025vcc} is held-out target-gene
generalization: train on one set of perturbations and evaluate on disjoint target
genes, ideally in a new biological screen. The input is an observed post-perturbation
expression profile and the target a scalar summary of its effect strength, so the task
is to generalize that mapping to unseen target genes, not to forecast a response from
perturbation identity alone. Strong performance within a single screen is therefore
not sufficient evidence of a signal that transfers.

\paragraph{Foundation models and the evaluation problem.}
Single-cell foundation models such as scGPT \citep{cui2024scgpt} and Geneformer
\citep{theodoris2023geneformer} are attractive here because their pretrained
representations may encode transferable biological structure. But cross-screen
evaluation is easy to get wrong. A representation may be informative only within
the screen it is tuned on; the prediction endpoint distributed with different
screens may measure different quantities; and simple summaries of the
response---its overall magnitude---can correlate with nuisance or design factors
such as the number of cells assayed per perturbation. Any of these can make an
in-distribution success look like transferable signal.

\paragraph{Core question.}
We therefore ask not whether any single feature dominates, but: \emph{what
survives a locked cross-screen evaluation once in-distribution informativeness,
zero-shot transfer, and endpoint/sample-size effects are separated?} We study a
frozen Geneformer representation $\zvec$ on VCC and two external Replogle CRISPRi
screens \citep{replogle2022mapping}, under the locked protocol of
Sec.~\ref{sec:protocol}. We compare $\zvec$ against a dimension-matched random-feature control $C_1$ evaluated
under the same selection protocol, and against shuffled controls, add a predefined four-scalar magnitude
block $\magmaptilde$, and score external transfer against a harmonized
target-gene endpoint rebuilt from single-cell data so that source and target
measure the same quantity.

\paragraph{Findings.}
Primary: the frozen representation carries measurable in-distribution information,
resolved even against a compressed-expression comparator, yet does not transport to
either external screen; and the strong in-domain magnitude signal is entangled with
sampling depth, so that reading requires refinement. Secondary: magnitude augmentation
improves external correlation without rescuing transfer, and a count-adjusted
max-response correlate is positive but correlational. We report the external pattern
descriptively, without inferring a mechanism.

\paragraph{Contributions.}
The contribution is not a new architecture but an evaluation audit: by separating
in-domain informativeness, cross-screen transfer, endpoint construction, and a key
design factor, the locked protocol changes the scientific conclusion a less controlled
evaluation would support.
\begin{enumerate}[itemsep=2pt,topsep=2pt,leftmargin=*]
\item \textbf{A locked, pre-registered cross-screen evaluation protocol} for
perturbation-effect prediction, with selection, controls and amendments fixed in
advance of the results they govern (Sec.~\ref{sec:protocol}).
\item \textbf{A foundation-model transfer result.} A frozen Geneformer
representation carries measurable VCC in-distribution predictive information
beyond a dimension-matched control ($\Delta\rsq(\zvec-C_1) = +0.1645$, $95\%$ CI
$[+0.1375, +0.1920]$) yet fails zero-shot transfer on both external screens
($\rho = -0.139$ and $-0.267$), lying below that control---in-domain information
is not evidence of portability.
\item \textbf{An endpoint / sampling-depth diagnosis.} The VCC endpoint is
strongly sample-size associated, and a count-only linear model out-predicts the
four magnitude scalars in validation ($\rsq = +0.4325$ vs.\ $+0.2589$, magnitude
adding $+0.0017$ beyond count), so the aggregate-magnitude interpretation is
substantially entangled with cell count.
\item \textbf{Magnitude augmentation, characterized precisely.} Adding the
predefined magnitude block improves the frozen representation externally
($\Delta\rho = +0.032$ and $+0.143$) but, under the frozen primary head, does not rescue
transfer (both remain negative).
\end{enumerate}

\section{Related work}
\label{sec:related}

\paragraph{Perturbation prediction.}
\spsm{}-style stable-prediction models descend from stable-prediction
methods in causal inference \citep{kuang2018stableprediction,
shen2018causally} adapted to single-cell data. Recent perturbation
prediction methods include CPA \citep{lotfollahi2023predicting}, GEARS
\citep{roohani2024gears}, and pretrained foundation models such as scGPT
\citep{cui2024scgpt}, Universal Cell Embeddings
\citep{rosen2026uce}, and Geneformer
\citep{theodoris2023geneformer}. Many existing evaluations either pool across
target-gene splits or use a held-out-pair setting; the strict disjoint
target-gene split used here is more demanding, and our analysis is
complementary: rather than proposing a new architecture we ask which
features any model class needs to exploit on this split.

\paragraph{Linear baselines and mode collapse in perturbation prediction.}
A recent line of work documents that deep and foundation-model
perturbation predictors frequently fail to outperform simple
baselines. \citet{ahlmann2025linear} compared five foundation models
and two deep architectures against linear baselines on Adamson,
Norman, and Replogle, and found that none consistently outperformed
the linear model on unseen-perturbation prediction; for held-out
perturbations the deep models often failed to outperform predicting
the unconditional training mean. They additionally show that a linear
model with perturbation embeddings pretrained on Replogle K562
transfers across cell lines to RPE1 (and vice versa), establishing
that cross-cell-line transfer with linear models is feasible.
\citet{wu2025perturbench} introduced PerturBench with covariate-transfer
benchmarks across cell lines on Srivatsan20 and Jiang24, and identified
mode collapse---predictions concentrating in a narrow region of output
space---as a recurring failure mode, with simple baselines (Latent
Additive, Decoder-Only) often matching or exceeding more complex
autoencoder models. \citet{csendes2025benchmarking} found that a mean baseline exceeded scGPT and scFoundation~\citep{hao2024scfoundation} on post-perturbation
Pearson($\Delta$), and \citet{bendidi2024onepca} reported that
PCA-based baselines remained competitive with transcriptomics
foundation models. \citet{mejia2025diversity} addressed mode collapse
on the loss side with a DEG-weighted training objective. Our work sits in this lineage
but differs in what it controls: we isolate response magnitude with pre-specified shuffle
and Gaussian controls (App.~\ref{sec:causal}), evaluate cross-screen transfer of a frozen
foundation-model representation under a locked protocol (Sec.~\ref{sec:transfer}), and
show that the endpoint distributed with those screens measures a different quantity from
the training target, rebuilding a common one (Sec.~\ref{sec:endpoint-problem}).
The evaluation uses the held-out
target-gene split of the Virtual Cell Challenge \citep{roohani2025vcc},
distinct from the covariate-transfer splits of \citet{wu2025perturbench}
and the unseen-perturbation splits of \citet{ahlmann2025linear}.

\paragraph{Imbalanced regression and tail prediction.}
Deep imbalanced regression frameworks include Label Distribution
Smoothing (LDS) and Feature Distribution Smoothing
\citep{yang2021delving}, Balanced MSE \citep{ren2022balancedmse},
density-based reweighting \citep{steininger2021densityrebalancing}, and
label-distribution-aware margin losses \citep{cao2019ldam}. We
instantiate LDS-weighted and tail-weighted variants as baselines
(App.~\ref{sec:diagnostic}); both fail in our setting, and
App.~\ref{app:additional-discussion} sets out how we read that failure.

\paragraph{Distribution shift and rank objectives.}
Cell-type transfer is a distribution shift between domains with
distinct gene-expression baselines. Standard OOD techniques such as IRM
\citep{arjovsky2019irm}, GroupDRO \citep{sagawa2020groupdro}, REx
\citep{krueger2021rex}, and the WILDS benchmark family
\citep{koh2021wilds} address related but distinct shifts and are
complementary. Our use of magnitude as a deliberately low-dimensional,
coordinate-permutation-invariant summary is conceptually allied with
norm- and energy-based detection \citep{hendrycks2017baseline,
liu2020energy} and norm-aware embeddings \citep{wang2017normface}. The
\magrank{} extension (App.~\ref{app:rank}) uses a margin-based pairwise
rank loss in the spirit of LambdaRank-family approaches
\citep{burges2005learning, cao2007listnet, hadsell2006dimensionality}.

\section{Problem setup and notation}
\label{sec:setup}

\paragraph{Data.}
The Virtual Cell Challenge (VCC) corpus \citep{roohani2025vcc} consists
of CRISPRi (dCas9--KRAB) \citep{gilbert2014crispri} Perturb-seq
\citep{dixit2016perturbseq} screens in H1 human embryonic stem cells
(H1 hESC), organized into 48 experimental batches indexed by
$b \in \mathcal{B}$. Each row $i$ corresponds to a (batch, target-gene)
pair $(b_i, g_i)$ with feature vector $\xvec_i \in \Rb^{d}$ and scalar
label $y_i \in \Rb_{\geq 0}$. Features are log1p-normalized mean
expression deltas relative to the NT-control mean of batch $b_i$,
restricted to the $d=2{,}000$ highly variable genes (HVGs) selected on
the train split. The label is $y_i = \log(1 + \max\{0,\, A^2_{b_i, g_i}\})$,
where $A^2_{b, g}$ is the two-sample Anderson--Darling statistic
\citep{scholz1987k} between the log1p expression of gene $g$ in
cells perturbed at $g$ within batch $b$ and in non-targeting
controls from the same batch; the AD statistic is computed on the
target-gene column only, and $\xvec_i$ enters as predictor input only
(App.~\ref{app:ad-label}).
The set of target genes is partitioned disjointly into
$\mathcal{G}_{\text{tr}}$, $\mathcal{G}_{\text{val}}$, and
$\mathcal{G}_{\text{te}}$ of sizes $140$, $49$, and $93$, giving
$5{,}740$ training, $2{,}052$ validation, and $3{,}972$ test rows. These are counts
after eligibility, which requires at least ten cells in some (target gene, batch) group
and removes $10$, $1$, and $7$ genes from the $150$, $50$, and $100$ raw targets; every
excluded gene is accounted for by that rule, with none left over.

\paragraph{Task and metrics.}
Given $(b_i, \xvec_i)$ for a test row, predict $y_i$. We report pooled
test
\(
\rsq = 1 - \sum_i (y_i - \hat{y}_i)^2 \big/ \sum_i (y_i -
\bar{y}_{\text{te}})^2,
\)
seed-level mean $\pm$ standard deviation across independent
training runs (default: seeds $\{0,1,2\}$; MAG-Anchor uses five seeds), Spearman rank
correlation, and per-quintile bias $\Eb[\hat{y} - y \mid q(y)]$. Standard deviations are
sample standard deviations throughout.

\paragraph{Cross-dataset shift.}
For external transfer we use two Replogle CRISPRi (dCas9--KRAB)
Perturb-seq screens \citep{replogle2022mapping, replogle2022dataset}:
RPE1 and K562-essential.
Features are mapped into the VCC HVG namespace by gene symbol; coverage is
screen-specific---K562-essential recovers $1{,}430$ of $2{,}000$ HVGs
($71.5\%$) and RPE1 recovers $1{,}550$ of $2{,}000$ ($77.5\%$); unmapped coordinates
are imputed as zero, the value an unperturbed cell takes in this pipeline. For each
screen we rebuild the label from single-cell data using the definition
above, so that source and target are scored on the same quantity---the
Anderson--Darling column distributed with the Replogle release measures
something else (Sec.~\ref{sec:endpoint-problem}). The rebuilt endpoint fixes subsample
sizes at the VCC medians so the statistic is not driven by sample-size differences, and
eligibility requires those cell counts and an unambiguous target mapping, a threshold
whose residual selection pathway we return to in Sec.~\ref{sec:discussion}. We never
train on either screen and never use target-side statistics for preprocessing; a legacy
training control on the shared-HVG subspace is reported in App.~\ref{app:hvg-control}.

\section{Locked evaluation protocol}
\label{sec:protocol}

Whether a representation \emph{transfers} is easy to overstate: an
in-distribution success, a control too weak to be informative, or an outcome
inspected after the fact can each inflate an apparent transfer signal. We
therefore fix each analysis-governing choice before the result it governs and
interpret transfer only through that recorded protocol (Fig.~\ref{fig:protocol}).

\paragraph{Pre-registration and freezing.}
Hypotheses, controls, metrics, the resampling procedures
(App.~\ref{app:uncertainty}) and decision rules are recorded in advance. Here,
``pre-registered'' denotes an internal pre-analysis specification and lock rather
than registration in an external registry; App.~\ref{app:chronology-evidence}
gives the chronology evidence and its limitations. Model and head selection use
VCC training and validation only and are frozen before any test metric is
computed; the locked protocol required external outcome labels to be withheld until
final unblinding. Selection never uses a test or external metric.

\paragraph{An informativeness gate before transfer.}
Interpretation is staged. The in-distribution gate (H-FM1) asks whether $\zvec$
carries usable signal on VCC \emph{beyond a dimension-matched random control}
$C_1$, matched in input dimension and in protocol (App.~\ref{app:controls}): it is
informative iff the paired bootstrap interval on
$\Delta\rsq = \rsq(\text{FM-only}) - \rsq(C_1)$ excludes zero on the positive
side, with $C_1$ alone as the gate. Zero-shot transfer (H-FM2) is interpreted
\emph{only if} that gate passes; if it fails, a near-zero external correlation
cannot be read as a transfer result (App.~\ref{app:hypotheses}). Passing the gate
does not predict that transfer succeeds.

\paragraph{Controls, primaries, and what stays open.}
Fixed controls---the dimension-matched Gaussian $C_1$ and shuffled-alignment
controls---are generated from array shapes only, never from labels. Two results
are primary: in-distribution informativeness together with the zero-shot transfer
outcome, and the endpoint/sampling-depth diagnosis. Magnitude augmentation and a
count-adjusted max-response summary are secondary. Hypotheses left unresolved
under the lock---including test-level magnitude recoverability (H-FM4)---are
reported as unresolved. All four hypotheses, their gating controls and their outcomes are in
App.~\ref{app:hypotheses}.

\section{Baseline failure and the magnitude construction}
\label{sec:magfeat}

\paragraph{Why magnitude.}
The \spsm{} V4 baseline, a stable-prediction MLP encoder with a within-context
stability loss and a relation loss (App.~\ref{app:spsm}), attains test pooled
$\rsq = +0.0824 \pm 0.0100$ across three seeds and exhibits \emph{mean collapse}:
its predictions concentrate near the marginal training mean, with monotonic
per-quintile bias that over-predicts the lowest quintile and under-predicts the
highest. We say ``mean collapse'' rather than ``mode collapse''
\citep{wu2025perturbench}, since predictions concentrate at one point rather than
at several modes. A pre-specified diagnostic suite showed that one-dimensional
magnitude summaries separate low- from high-response rows about as well as the
full $2{,}000$-dimensional expression vector, which motivates supplying response
\emph{magnitude} to the predictor as an explicit feature rather than through the
loss. Earlier exploratory work tried five loss- and pretraining-side corrections
that failed or stalled; those runs predate the locked evaluation and their
artifacts were not retained, so we use them as historical motivation only and draw
no quantitative or mechanistic conclusion from them.
The diagnostic suite, the proof-of-concept encoder \magspsm{}, the MAG-Anchor
variant and the per-row alignment argument are in App.~\ref{app:why-hard}.

\paragraph{The magnitude block.}
Two later results (Sec.~\ref{sec:magnitude}, Sec.~\ref{sec:max-response}) are
stated in terms of this block, so we define it here. For $\xvec \in \Rb^{d}$ the magnitude map $\magmap : \Rb^{d} \to \Rb^{4}$ collects
four scalars, $m_1 = \tfrac{1}{d}\sum_j |x_j|$, $m_2 = \|\xvec\|_2$,
$m_3 = \max_j |x_j|$ and $m_4 = \sigma(\xvec)$, each $z$-scored component-wise
using \emph{training-set} statistics to give $\magmaptilde(\xvec)$
(Eq.~\ref{eq:mag-features}); standardization removes any constant factor, so
specifying $m_2$ as $\|\xvec\|_2$ or as $\|\xvec\|_2/\sqrt{d}$ gives an identical
model input. All four are coordinate-permutation invariant, $m_1$ to $m_3$ are
sign-flip invariant, and $m_4$ is invariant under translations along
$\mathbf{1}$. Those invariances are the point: the block summarizes \emph{how
much} a cell responded rather than \emph{which genes} did. That is what makes it
a candidate shift-robust summary across cell types, and equally what makes it
vulnerable to any nuisance factor that scales the response, as
Sec.~\ref{sec:endpoint-depth} shows sampling depth does.

\section{Experiments}
\label{sec:exp}

\paragraph{The frozen representation.}
The representation $\zvec$ is Geneformer-V2-104M
\citep{theodoris2023geneformer} used strictly as a frozen feature extractor: the
pretrained weights are loaded in inference mode and never updated, so no gradient
reaches the encoder anywhere in this study. Each cell is tokenized by the model's
own rank-value encoding and embedded to $768$ dimensions; a perturbation-level
vector is the mean embedding over the cells of a construct minus a non-targeting
reference mean, $\zvec_g = |g|^{-1}\sum_{i \in g} E_i - \mu_{\mathrm{NT}}$. The
non-targeting reference is batch-local for VCC and globally pooled per external
screen; that asymmetry, the pinned checkpoint digests, and the other recorded
cross-screen differences are set out in App.~\ref{app:fm-representation},
together with the standardization, which is fit on the VCC training split alone.

\paragraph{Controls, heads, and other models.}
$C_1$ is the capacity floor that licenses every comparison below: a
$768$-dimensional matrix of independent standard Gaussian entries drawn per row
from array shapes alone, matched to $\zvec$ in input dimension and passed through
the identical preprocessing, candidate-head pool, selection rule and training
budget. The remaining controls, the candidate heads, the selection rule and the
freeze are specified in App.~\ref{app:controls} and App.~\ref{app:heads}, and all
five controls were fixed in advance. The deep \spsm{} and \magspsm{} models share the
\spsm{} V4 backbone (App.~\ref{app:spsm}) and MAG-Anchor a two-branch
architecture (App.~\ref{app:mag-anchor-full}); their hyperparameters and compute
are in App.~\ref{app:spsm}, and those of the classical baselines in
App.~\ref{app:classical}. We report mean $\pm$ std
over seeds $\{0,1,2\}$ (MAG-Anchor uses five seeds).

\subsection{In-distribution informativeness}
\label{sec:informativeness}

Before interpreting any transfer result we verify that the frozen representation
carries usable in-distribution signal---the pre-registered informativeness gate
(H-FM1). On the VCC held-out target-gene test set, the FM-only arm $\zvec$ improves
pooled $\rsq$ over its capacity-matched Gaussian control $C_1$ by a resolved paired
margin, $\Delta\rsq(\zvec-C_1) = +0.1645$ ($95\%$ CI $[+0.1375, +0.1920]$,
sign-stable across all resamples; row-level paired bootstrap,
App.~\ref{app:uncertainty}). The interval excludes zero on the positive side, so
the gate passes with $C_1$ alone as the criterion: the frozen representation is
informative in-distribution beyond a dimension-matched control. The frozen VCC test
split was scored once (App.~\ref{app:test-ledger}).
The pre-specified compressed-expression benchmark $C_2$ is reported alongside and
non-gating, as the specification requires: $\Delta\rsq(\zvec-C_2) = +0.1107$
($95\%$ CI $[+0.0906, +0.1317]$). Absolute pooled $\rsq$ on
the frozen test split is $+0.1496$ for $\zvec$, $+0.0387$ for $C_2$ and $-0.0150$
for $C_1$.

Head selection ran independently per arm under one pre-registered rule, which
resolved to a random forest for $\zvec$ and to histogram gradient boosting for
$C_1$; the selected $C_1$ head attains $\rsq = -0.015$ on the frozen test split.
We report this head-family asymmetry explicitly and read $C_1$ as the
pre-registered random-feature floor rather than as a same-head comparator
(App.~\ref{app:heads}).

\subsection{Zero-shot transfer across screens}
\label{sec:transfer}

\paragraph{A common endpoint.}
\label{sec:endpoint-problem}
Transfer is scored only where source and target measure the same quantity. The
Anderson--Darling column distributed with Replogle measures the breadth of a
transcriptome-wide response, not target-gene effect strength, and we could not reproduce
it from the single-cell data under any construction we tested
(App.~\ref{app:endpoint-provenance}). We therefore score transfer against the harmonized
target-gene endpoint rebuilt for each external screen (Sec.~\ref{sec:setup}), with
subsample sizes fixed at the VCC medians so the statistic is not driven by sample-size
differences. Eligibility retains $1{,}932$ RPE1 and $1{,}903$ K562-essential constructs
for transfer evaluation, with exclusions dominated by target genes absent from the
feature namespace rather than by insufficient cells
(App.~\ref{app:harmonized-endpoint}). A small fraction of evaluated constructs target
genes that also occur among the $150$ raw VCC training targets---$43$ of $1{,}932$ on
RPE1 ($2.2\%$) and $33$ of $1{,}903$ on K562-essential ($1.7\%$). The primary retains all eligible constructs,
as pre-registered; a pre-specified sensitivity that removes them leaves every verdict
unchanged (App.~\ref{app:hypotheses}). Models trained only on VCC are applied zero-shot with no
target-side calibration; uncertainty is a paired target-gene cluster bootstrap
(App.~\ref{app:uncertainty}).

\paragraph{Transfer.}
Throughout, a comparison is \emph{resolved} when its $95\%$ interval excludes zero.
Applied zero-shot to both external screens, the frozen representation does not
transfer. Its rank correlation with the harmonized target-gene endpoint is
negative on both screens under the pre-specified interval criterion---Spearman $\rho = -0.139$ on RPE1 ($95\%$ CI
$[-0.185, -0.092]$) and $\rho = -0.267$ on K562-essential ($[-0.310, -0.224]$), both
intervals lying entirely below zero. The capacity-matched control $C_1$ is by
contrast compatible with null on both screens ($\rho = -0.034$ $[-0.078, +0.012]$ and
$+0.032$ $[-0.012, +0.077]$, each interval spanning zero), as is a shuffled-alignment
control that destroys the representation's per-row identity while preserving its
marginal ($\rho = +0.044$ $[-0.001, +0.088]$ and $-0.004$ $[-0.049, +0.041]$). The
pre-specified compressed-expression control $C_2$ is also negative with its $95\%$ interval excluding zero on
both screens: on RPE1, $\rho = -0.156$ ($95\%$ CI $[-0.202, -0.109]$) versus
$-0.139$ for $\zvec$, and on K562-essential, $\rho = -0.264$ ($[-0.306, -0.221]$)
versus $-0.267$. Because $C_2$ contains no foundation-model computation, the
external failure is not specific to the frozen Geneformer representation; this
comparison does not identify its mechanism. The
representation is therefore not merely near zero: paired against its own noise floor
the paired contrast is negative with its interval excluding zero ($\rho_{\zvec} - \rho_{C_1} = -0.106$ $[-0.168, -0.044]$ on
RPE1 and $-0.299$ $[-0.360, -0.238]$ on K562-essential). Because the in-distribution
gate passed first, this is interpretable as a transfer failure---in-domain
information that does not carry across screens---rather than an uninformative
representation. In-distribution informativeness (Sec.~\ref{sec:informativeness})
followed by this governed zero-shot outcome together constitute the paper's first
primary result. The complete locked per-arm external results and persisted paired contrasts are
in App.~\ref{app:locked-external}, Table~\ref{tab:fm_external_primary}.

\subsection{Magnitude augmentation}
\label{sec:magnitude}

Adding the predefined four-scalar magnitude block $\magmaptilde$ to the frozen
representation improves it externally by a resolved paired margin. The paired
increment $\Delta\rho([\zvec;\magmaptilde] - \zvec)$ is $+0.032$ on RPE1 ($95\%$ CI
$[+0.008, +0.056]$) and $+0.143$ on K562-essential ($[+0.111, +0.175]$)---positive
with intervals excluding zero on both screens. The increment also remains resolved
against both pre-registered controls, $C_4$ (row-shuffled magnitude) and $C_5$ (four Gaussian
increment features), supporting dependence on aligned per-row magnitude rather than added
dimensionality alone. The
improvement does not, however, rescue transfer: the augmented representation remains
negative with its reported intervals excluding zero on both screens ($\rho = -0.107$ and $-0.124$) and still below
the capacity-matched control. Adding magnitude makes the representation less
anti-correlated with the outcome; it does not make it transfer. That holds under the frozen
primary head and construction basis: the pre-specified secondary analyses show the
augmented-arm result is head- and feature-basis-sensitive. Neither alters the $\zvec$-only
transfer result, and neither was used to select the reported primary
(App.~\ref{app:secondaries}).

\subsection{The endpoint is entangled with sampling depth}
\label{sec:endpoint-depth}

\begin{figure}[t]
  \centering
  \includegraphics[width=0.9\linewidth]{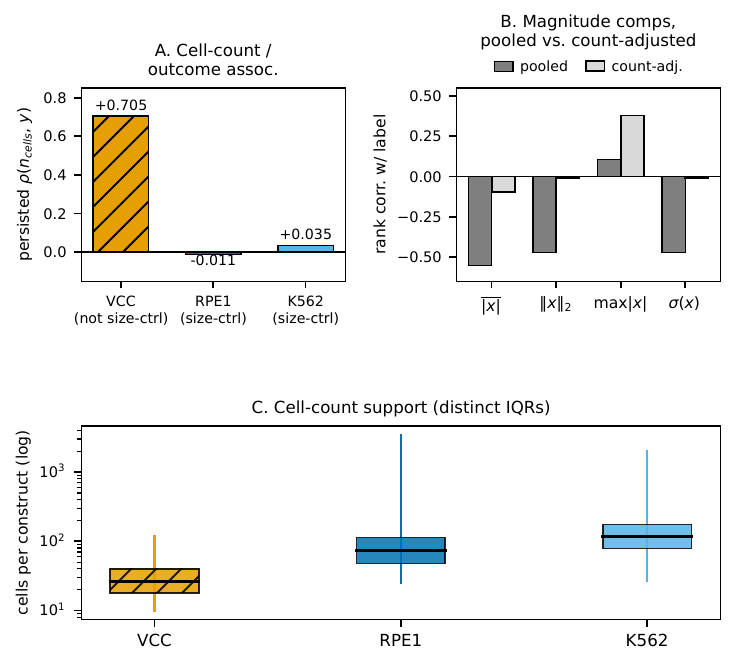}
  \caption{Sampling depth and the endpoint. \textbf{(A)}~Rank correlation between cells
  assayed per construct and the outcome: $+0.705$ on VCC, whose endpoint is not
  size-controlled, against $-0.011$ on RPE1 and $+0.035$ on K562-essential, whose
  harmonized endpoint is. \textbf{(B)}~The four magnitude summaries against the label,
  pooled and after adjusting for cell count; adjustment moves every summary toward zero
  except $\max_j|x_j|$, which strengthens. \textbf{(C)}~Cell-count support per construct on the checkpoint
  $\xvec_{\mathrm{cell}}$ accumulator population: the VCC interquartile range, $[18, 40]$
  cells, is disjoint from both external ranges, while RPE1, $[48, 113]$, and K562-essential,
  $[79, 174]$, overlap each other. App.~\ref{app:matched-sensitivity} uses the
  persisted-embedding set, on which the RPE1 summary differs.}
  \label{fig:sampling-depth}
\end{figure}

The second primary result concerns the benchmark endpoint itself. The VCC
Anderson--Darling label is strongly associated with the number of cells assayed per
perturbation (Spearman $\rho(n_{\text{cells}}, y) = +0.705$, Fig.~\ref{fig:sampling-depth}A):
unlike the harmonized external endpoint, the VCC endpoint is computed over all cells in a
group and is not size-controlled by construction. This association propagates into the magnitude features.
In VCC validation, a linear model on cell count alone reaches $\rsq = +0.4325$,
\emph{above} the four magnitude scalars' $\rsq = +0.2589$, and adding the four scalars to
cell count raises $\rsq$ by only $+0.0017$. Much of the
aggregate-magnitude signal that predicts this endpoint therefore overlaps with sampling
depth on this benchmark.

This is an association on this benchmark, not a causal claim: partialling out an observed
covariate does not establish a causal role, and it does not mean the label ``is'' cell
count---the two are strongly associated on VCC. Nor does it extend automatically to the
external screens, whose endpoint is size-controlled and where the association between
cell count and outcome is near zero on the eligible sets ($\rho = -0.011$, $p = 0.63$ on
RPE1; $\rho = +0.035$ on K562-essential)---though near zero is not proof of absence, and the
eligibility threshold excludes sub-threshold constructs, a residual selection pathway whose
pre-eligibility count is not recoverable from the record. The aggregate-magnitude reading is
therefore entangled with a design factor---exactly what a locked evaluation should surface.

Unlike the endpoint, the magnitude and representation blocks are not size-matched across
screens; under a pre-specified matched-count sensitivity the cross-screen ordering changes,
so that comparison is depth-dependent---a property of the comparison, not an explanation of
the transfer failure (App.~\ref{app:matched-sensitivity}).

\subsection{A count-adjusted max-response correlate}
\label{sec:max-response}

One pre-registered secondary survives count adjustment. After partialling out
$\log(1 + n_{\text{cells}})$, the single largest-magnitude coordinate $\max_j |x_j|$
retains a positive association with the outcome on VCC ($+0.379$) and on both external
screens ($+0.239$ on RPE1, $+0.355$ on K562-essential). We report point estimates only:
no interval procedure was pre-registered for this secondary and none is persisted, so we
do not attach uncertainty after the fact (App.~\ref{app:uncertainty}). The pre-registered
robustness check that \emph{was} specified---stratifying by
cell count---finds the association positive within every $n_{\text{cells}}$ quintile on both
external screens. We report this as a narrow, correlational, count-adjusted secondary:
positive on both screens, but not a recovered magnitude signal, not a new primary finding,
and not evidence that magnitude successfully transfers.

\subsection{What the locked evaluation establishes}
\label{sec:synthesis}

Read together, the results form one argument: in-distribution information does not imply
portability, and the benchmark endpoint that motivated the magnitude reading is itself
entangled with a design factor. What the protocol does not settle it leaves open, and we
say so rather than filling the gap post-hoc (Sec.~\ref{sec:discussion}). The contribution is the locked evaluation
itself, which separates in-distribution informativeness, transferable signal, endpoint
construction, and design factors, and thereby changes the conclusion a less controlled
analysis would reach.

\section{Discussion and limitations}
\label{sec:discussion}

\paragraph{What the transfer failure means.}
The frozen Geneformer representation is informative on VCC yet fails on both
external screens, below its capacity-matched control. In-domain predictive
information is therefore not sufficient evidence of cross-screen portability, and
strong held-out performance within one screen should not be read as transferable
signal. This is a statement about this representation under this evaluation, not
a claim that foundation models are generally unhelpful.

\paragraph{Why the magnitude interpretation changed, and what survives it.}
The count-only comparison of Sec.~\ref{sec:endpoint-depth} refines the earlier
aggregate-magnitude reading: the block is entangled with sampling depth rather than a clean
measure of response strength. What survives is narrower---the pre-registered,
count-adjusted max-response correlate of Sec.~\ref{sec:max-response}, which we do not
promote to a primary result.

\paragraph{Residual selection, and comparator sensitivity.}
Both caveats of Sec.~\ref{sec:endpoint-depth} stand: the external null is not proof of
absence, and the eligibility threshold's residual selection pathway is not quantifiable from
the record. Separately, the primary and secondary heads can differ materially, as can the
two external feature comparators; we fixed both in advance rather than selecting them after
seeing outcomes.

\paragraph{The pre-registered interval on VCC is unclustered.}
The in-distribution interval resamples rows, whereas the external intervals resample
target-gene clusters. Because VCC rows can share a target gene or a batch, the row-level
procedure may understate uncertainty relative to a clustered alternative; we retained the
pre-specified analysis rather than substituting one after the test evaluation.

\paragraph{What was not tested.}
One explanation is available but untested: $\zvec$ lands \emph{below} its
capacity-matched control externally rather than at zero, which could arise if
$\zvec$ partly encodes sampling depth while the VCC label is depth-associated and
the external one is not. We did not test it, deliberately: any analysis designed
after unblinding would be post-hoc (a later post-unblinding probe of linear depth recovery,
App.~\ref{app:post-unblinding}, settles nothing here). The pre-specified $C_2$ comparator
bears on this question; we report it descriptively and infer no mechanism from it. Magnitude
recoverability (H-FM4) was evaluated on train/validation only; its test-level statement
remains unresolved.

\paragraph{Lesson.}
The separation this evaluation enforces---in-domain informativeness, transferable signal,
endpoint construction, design factors like sampling depth, and primary versus secondary
evidence---is what changes the conclusion (Sec.~\ref{sec:synthesis}).

\bibliographystyle{plainnat}
\bibliography{references}

@article{replogle2022mapping,
  title   = {Mapping information-rich genotype-phenotype landscapes with
             genome-scale {Perturb-seq}},
  author  = {Replogle, Joseph M. and Saunders, Reuben A. and Pogson,
             Angela N. and Hussmann, Jeffrey A. and Lenail, Alexander and
             Guna, Alina and Mascibroda, Lauren and Wagner, Eric J. and
             Adelman, Karen and Lithwick-Yanai, Gila and Iremadze,
             Nika and Oberstrass, Florian and Lipson, Doron and
             Bonnar, Jessica L. and Jost, Marco and Norman, Thomas M.
             and Weissman, Jonathan S.},
  journal = {Cell},
  volume  = {185},
  number  = {14},
  pages   = {2559--2575.e28},
  year    = {2022},
  doi     = {10.1016/j.cell.2022.05.013}
}

@misc{replogle2022dataset,
  title        = {{``Mapping information-rich genotype-phenotype
                  landscapes with genome-scale Perturb-seq''} Replogle
                  et al.\ 2022 processed {Perturb-seq} datasets},
  author       = {Replogle, Joseph M. and Weissman, Jonathan S.},
  year         = {2022},
  publisher    = {Figshare+},
  doi          = {10.25452/figshare.plus.20029387},
  note         = {Includes K562 genome-scale, K562 essential, and RPE1
                  essential Perturb-seq AnnData files, in pseudo-bulk and
                  cell-level form. We use the RPE1 and K562-essential
                  pseudo-bulk and raw single-cell releases.}
}

@article{dixit2016perturbseq,
  title   = {Perturb-Seq: Dissecting Molecular Circuits with Scalable
             Single-Cell RNA Profiling of Pooled Genetic Screens},
  author  = {Dixit, Atray and Parnas, Oren and Li, Biyu and Chen, Jenny
             and Fulco, Charles P. and Jerby-Arnon, Livnat and
             Marjanovic, Nemanja D. and Dionne, Danielle and Burks,
             Tyler and Raychowdhury, Raktima and others},
  journal = {Cell},
  volume  = {167},
  number  = {7},
  pages   = {1853--1866.e17},
  year    = {2016},
  doi     = {10.1016/j.cell.2016.11.038}
}

@article{lotfollahi2019scgen,
  title   = {scGen predicts single-cell perturbation responses},
  author  = {Lotfollahi, Mohammad and Wolf, F. Alexander and Theis, Fabian J.},
  journal = {Nature Methods},
  volume  = {16},
  number  = {8},
  pages   = {715--721},
  year    = {2019},
  doi     = {10.1038/s41592-019-0494-8}
}

@article{lotfollahi2023predicting,
  title   = {Predicting cellular responses to complex perturbations in
             high-throughput screens},
  author  = {Lotfollahi, Mohammad and Klimovskaia Susmelj, Anna and De
             Donno, Carlo and Hetzel, Leon and Ji, Yuge and Ibarra,
             Ignacio L. and Srivatsan, Sanjay R. and Naghipourfar, Mohsen
             and Daza, Riza M. and Martin, Beth and others},
  journal = {Molecular Systems Biology},
  volume  = {19},
  number  = {6},
  pages   = {e11517},
  year    = {2023},
  doi     = {10.15252/msb.202211517}
}

@article{roohani2024gears,
  title   = {Predicting transcriptional outcomes of novel multigene
             perturbations with GEARS},
  author  = {Roohani, Yusuf and Huang, Kexin and Leskovec, Jure},
  journal = {Nature Biotechnology},
  volume  = {42},
  number  = {6},
  pages   = {927--935},
  year    = {2024},
  doi     = {10.1038/s41587-023-01905-6}
}

@article{cui2024scgpt,
  title   = {scGPT: toward building a foundation model for single-cell
             multi-omics using generative AI},
  author  = {Cui, Haotian and Wang, Chloe and Maan, Hassaan and Pang,
             Kuan and Luo, Fengning and Duan, Nan and Wang, Bo},
  journal = {Nature Methods},
  volume  = {21},
  number  = {8},
  pages   = {1470--1480},
  year    = {2024},
  doi     = {10.1038/s41592-024-02201-0}
}

@article{theodoris2023geneformer,
  title   = {Transfer learning enables predictions in network biology},
  author  = {Theodoris, Christina V. and Xiao, Ling and Chopra, Anant
             and Chaffin, Mark D. and Al Sayed, Zeina R. and Hill,
             Matthew C. and Mantineo, Helene and Brydon, Elizabeth M.
             and Zeng, Zexian and Liu, X. Shirley and Ellinor, Patrick
             T.},
  journal = {Nature},
  volume  = {618},
  number  = {7965},
  pages   = {616--624},
  year    = {2023},
  doi     = {10.1038/s41586-023-06139-9}
}

@article{rosen2026uce,
  title   = {Universal cell embedding provides a foundation model for
             cell biology},
  author  = {Rosen, Yanay and Roohani, Yusuf and Agrawal, Ayush and
             Samotor{\v{c}}an, Leon and {Tabula Sapiens Consortium} and
             Quake, Stephen R. and Leskovec, Jure},
  journal = {Nature},
  volume  = {656},
  number  = {8126},
  pages   = {183--191},
  year    = {2026},
  doi     = {10.1038/s41586-026-10689-z}
}

@inproceedings{koh2021wilds,
  title   = {WILDS: A Benchmark of in-the-Wild Distribution Shifts},
  author  = {Koh, Pang Wei and Sagawa, Shiori and Marklund, Henrik and
             Xie, Sang Michael and Zhang, Marvin and Balsubramani,
             Akshay and Hu, Weihua and Yasunaga, Michihiro and Phillips,
             Richard Lanas and Gao, Irena and others},
  booktitle = {Proceedings of the 38th International Conference on
               Machine Learning (ICML)},
  series    = {Proceedings of Machine Learning Research},
  volume    = {139},
  pages     = {5637--5664},
  publisher = {PMLR},
  year      = {2021}
}

@article{arjovsky2019irm,
  title   = {Invariant Risk Minimization},
  author  = {Arjovsky, Mart{\'i}n and Bottou, L{\'e}on and Gulrajani,
             Ishaan and Lopez-Paz, David},
  journal = {arXiv preprint arXiv:1907.02893},
  year    = {2019}
}

@inproceedings{sagawa2020groupdro,
  title   = {Distributionally Robust Neural Networks for Group Shifts:
             On the Importance of Regularization for Worst-Case
             Generalization},
  author  = {Sagawa, Shiori and Koh, Pang Wei and Hashimoto, Tatsunori
             B. and Liang, Percy},
  booktitle = {International Conference on Learning Representations (ICLR)},
  year      = {2020}
}

@inproceedings{krueger2021rex,
  title   = {Out-of-Distribution Generalization via Risk Extrapolation
             (REx)},
  author  = {Krueger, David and Caballero, Ethan and Jacobsen, J{\"o}rn-
             Henrik and Zhang, Amy and Binas, Jonathan and Zhang,
             Dinghuai and Le Priol, R{\'e}mi and Courville, Aaron},
  booktitle = {International Conference on Machine Learning (ICML)},
  series    = {Proceedings of Machine Learning Research},
  volume    = {139},
  pages     = {5815--5826},
  publisher = {PMLR},
  year      = {2021}
}

@inproceedings{hendrycks2017baseline,
  title   = {A Baseline for Detecting Misclassified and
             Out-of-Distribution Examples in Neural Networks},
  author  = {Hendrycks, Dan and Gimpel, Kevin},
  booktitle = {International Conference on Learning Representations (ICLR)},
  year      = {2017}
}

@inproceedings{liu2020energy,
  title   = {Energy-based Out-of-Distribution Detection},
  author  = {Liu, Weitang and Wang, Xiaoyun and Owens, John D. and Li,
             Yixuan},
  booktitle = {Advances in Neural Information Processing Systems (NeurIPS)},
  volume    = {33},
  pages     = {21464--21475},
  year      = {2020}
}

@inproceedings{wang2017normface,
  title   = {NormFace: $L_2$ Hypersphere Embedding for Face Verification},
  author  = {Wang, Feng and Xiang, Xiang and Cheng, Jian and Yuille, Alan L.},
  booktitle = {Proceedings of the 25th ACM International Conference on
               Multimedia},
  pages     = {1041--1049},
  year      = {2017},
  doi       = {10.1145/3123266.3123359}
}

@inproceedings{yang2021delving,
  title     = {Delving into Deep Imbalanced Regression},
  author    = {Yang, Yuzhe and Zha, Kaiwen and Chen, Ying-Cong and Wang,
               Hao and Katabi, Dina},
  booktitle = {International Conference on Machine Learning (ICML)},
  series    = {Proceedings of Machine Learning Research},
  volume    = {139},
  pages     = {11842--11851},
  publisher = {PMLR},
  year      = {2021}
}

@inproceedings{ren2022balancedmse,
  title   = {Balanced MSE for Imbalanced Visual Regression},
  author  = {Ren, Jiawei and Zhang, Mingyuan and Yu, Cunjun and Liu, Ziwei},
  booktitle = {IEEE/CVF Conference on Computer Vision and Pattern
               Recognition (CVPR)},
  pages     = {7926--7935},
  year      = {2022}
}

@article{steininger2021densityrebalancing,
  title   = {Density-based weighting for imbalanced regression},
  author  = {Steininger, Michael and Kobs, Konstantin and Davidson, Padraig
             and Krause, Anna and Hotho, Andreas},
  journal = {Machine Learning},
  volume  = {110},
  number  = {8},
  pages   = {2187--2211},
  year    = {2021},
  doi     = {10.1007/s10994-021-06023-5}
}

@inproceedings{cao2019ldam,
  title   = {Learning Imbalanced Datasets with Label-Distribution-Aware
             Margin Loss},
  author  = {Cao, Kaidi and Wei, Colin and Gaidon, Adrien and Ar{\'e}chiga,
             Nikos and Ma, Tengyu},
  booktitle = {Advances in Neural Information Processing Systems (NeurIPS)},
  volume    = {32},
  pages     = {1565--1576},
  year      = {2019}
}

@inproceedings{burges2005learning,
  title   = {Learning to Rank using Gradient Descent},
  author  = {Burges, Chris and Shaked, Tal and Renshaw, Erin and Lazier,
             Ari and Deeds, Matt and Hamilton, Nicole and Hullender,
             Greg},
  booktitle = {Proceedings of the 22nd International Conference on Machine
               Learning (ICML)},
  pages     = {89--96},
  year      = {2005},
  doi       = {10.1145/1102351.1102363}
}

@inproceedings{cao2007listnet,
  title   = {Learning to rank: from pairwise approach to listwise
             approach},
  author  = {Cao, Zhe and Qin, Tao and Liu, Tie-Yan and Tsai, Ming-Feng
             and Li, Hang},
  booktitle = {Proceedings of the 24th International Conference on Machine
               Learning (ICML)},
  pages     = {129--136},
  year      = {2007},
  doi       = {10.1145/1273496.1273513}
}

@inproceedings{hadsell2006dimensionality,
  title   = {Dimensionality Reduction by Learning an Invariant Mapping},
  author  = {Hadsell, Raia and Chopra, Sumit and LeCun, Yann},
  booktitle = {IEEE Conference on Computer Vision and Pattern Recognition
               (CVPR)},
  volume    = {2},
  pages     = {1735--1742},
  year      = {2006},
  doi       = {10.1109/CVPR.2006.100}
}

@inproceedings{kuang2018stableprediction,
  title     = {Stable Prediction across Unknown Environments},
  author    = {Kuang, Kun and Cui, Peng and Athey, Susan and Xiong, Ruoxuan
               and Li, Bo},
  booktitle = {Proceedings of the 24th ACM SIGKDD International Conference
               on Knowledge Discovery \& Data Mining},
  pages     = {1617--1626},
  year      = {2018},
  doi       = {10.1145/3219819.3220082}
}

@inproceedings{shen2018causally,
  title     = {Causally Regularized Learning with Agnostic Data Selection
               Bias},
  author    = {Shen, Zheyan and Cui, Peng and Kuang, Kun and Li, Bo and
               Chen, Peixuan},
  booktitle = {Proceedings of the 26th ACM International Conference on
               Multimedia},
  pages     = {411--419},
  year      = {2018},
  doi       = {10.1145/3240508.3240577}
}

@article{scholz1987k,
  title   = {K-Sample Anderson-Darling Tests},
  author  = {Scholz, F. W. and Stephens, M. A.},
  journal = {Journal of the American Statistical Association},
  volume  = {82},
  number  = {399},
  pages   = {918--924},
  year    = {1987},
  doi     = {10.1080/01621459.1987.10478517}
}

@article{roohani2025vcc,
  title       = {Virtual Cell Challenge: Toward a {Turing} test for the virtual cell},
  author      = {Roohani, Yusuf H. and Hua, Tony J. and Tung, Po-Yuan and
                 Bounds, Lexi R. and Yu, Feiqiao B. and Dobin, Alexander and
                 Teyssier, Noam and Adduri, Abhinav and Woodrow, Alden and
                 Plosky, Brian S. and Mehta, Reshma and Hsu, Benjamin and
                 Sullivan, Jeremy and Ricci-Tam, Chiara and Li, Nianzhen and
                 Kazaks, Julia and Gilbert, Luke A. and Konermann, Silvana and
                 Hsu, Patrick D. and Goodarzi, Hani and Burke, Dave P.},
  journal     = {Cell},
  volume      = {188},
  number      = {13},
  pages       = {3370--3374},
  year        = {2025},
  doi         = {10.1016/j.cell.2025.06.008},
  publisher   = {Elsevier}
}

@article{gilbert2014crispri,
  title     = {Genome-scale {CRISPR}-mediated control of gene repression and activation},
  author    = {Gilbert, Luke A. and Horlbeck, Max A. and Adamson, Britt and
               Villalta, Jacqueline E. and Chen, Yuwen and Whitehead, Evan H. and
               Guimaraes, Carla and Panning, Barbara and Ploegh, Hidde L. and
               Bassik, Michael C. and Qi, Lei S. and Kampmann, Martin and
               Weissman, Jonathan S.},
  journal   = {Cell},
  volume    = {159},
  number    = {3},
  pages     = {647--661},
  year      = {2014},
  doi       = {10.1016/j.cell.2014.09.029},
  publisher = {Elsevier}
}

@article{ahlmann2025linear,
  author  = {Ahlmann-Eltze, Constantin and Huber, Wolfgang and Anders, Simon},
  title   = {Deep-learning-based gene perturbation effect prediction does not yet outperform simple linear baselines},
  journal = {Nature Methods},
  volume  = {22},
  number  = {8},
  pages   = {1657--1661},
  year    = {2025},
  doi     = {10.1038/s41592-025-02772-6}
}

@inproceedings{wu2025perturbench,
  author    = {Wu, Yan and Wershof, Esther and Schmon, Sebastian M. and Nassar, Marcel and Osi{\'n}ski, B{\l}a{\.z}ej and Eksi, Ridvan and Yan, Zichao and Stark, Rory and Zhang, Kun and Graepel, Thore},
  title     = {{PerturBench}: Benchmarking Machine Learning Models for Cellular Perturbation Analysis},
  booktitle = {Advances in Neural Information Processing Systems Datasets and Benchmarks Track},
  year      = {2025}
}

@article{csendes2025benchmarking,
  author  = {Csendes, Gerold and Sanz, Gema and Szalay, Krist{\'o}f Z. and Szalai, Bence},
  title   = {Benchmarking foundation cell models for post-perturbation {RNA-seq} prediction},
  journal = {BMC Genomics},
  volume  = {26},
  number  = {1},
  pages   = {393},
  year    = {2025},
  doi     = {10.1186/s12864-025-11600-2}
}

@inproceedings{bendidi2024onepca,
  author    = {Bendidi, Ihab and Whitfield, Shawn and Kenyon-Dean, Kian and Ben Yedder, Hanene and El Mesbahi, Yassir and Noutahi, Emmanuel and Denton, Alisandra K.},
  title     = {Benchmarking Transcriptomics Foundation Models for Perturbation Analysis: One {PCA} Still Rules Them All},
  booktitle = {NeurIPS Workshop on AI for New Drug Modalities (AIDrugX)},
  year      = {2024}
}

@inproceedings{mejia2025diversity,
  author    = {Mejia, Gabriel M. and Miller, Henry E. and Leblanc, Francis J. A. and Wang, Bo and Swain, Brendan and de Lima Camillo, Lucas Paulo},
  title     = {Diversity by Design: Addressing Mode Collapse Improves {scRNA-seq} Perturbation Modeling on Well-Calibrated Metrics},
  booktitle = {2nd Generative AI for Biology Workshop at the 42nd
               International Conference on Machine Learning (ICML)},
  year      = {2025},
  note      = {Non-archival workshop paper. Preprint {arXiv:2506.22641}}
}

@article{hao2024scfoundation,
  title   = {Large-scale foundation model on single-cell transcriptomics},
  author  = {Hao, Minsheng and Gong, Jing and Zeng, Xin and Liu, Chiming
             and Guo, Yucheng and Cheng, Xingyi and Wang, Taifeng and
             Ma, Jianzhu and Zhang, Xuegong and Song, Le},
  journal = {Nature Methods},
  year    = {2024},
  volume  = {21},
  number  = {8},
  pages   = {1481--1491},
  doi     = {10.1038/s41592-024-02305-7}
}

\subsection*{AI use statement}

Generative AI assistants were used to help draft the manuscript and to write and review
analysis code. Assistants were not used to train or refit predictive models, or to
re-evaluate the held-out test split. The research question, the decisions taken at each
stage of the protocol, and the interpretation of every result are the authors', who
reviewed all assistant-produced material before adoption and take full responsibility for
the content of this paper, including any errors it contains.

\subsection*{Ethics statement}
This work uses publicly released CRISPRi Perturb-seq screens of cultured
cell lines---the Virtual Cell Challenge release and the Replogle RPE1 and
K562-essential datasets. It involves no human subjects, no personally
identifying information, and no new data collection. We introduce no
deployed model or clinical decision system; the study is a methodological
diagnostic on public benchmarks, and any societal effects would be
indirect and mediated through future perturbation-prediction methods.

\subsection*{Reproducibility statement}
The evaluation is governed by a locked, pre-registered protocol
(Section~\ref{sec:protocol}): heads and model selection are frozen on VCC
before any test metric is computed, the locked protocol required the external outcome labels
to be held out until final evaluation, and analysis-governing choices are fixed
before the results they govern, with subsequent amendments recorded as dated
documents. Controls, metrics, the paired target-gene
cluster bootstrap, and decision rules are specified in advance; the
problem setup, endpoint construction, magnitude features, seeds, and
hyperparameters are given in Sections~\ref{sec:setup}--\ref{sec:exp} and
the appendices. As disclosed in App.~\ref{app:legacy-not-rederived}, the
classical baseline tables from an earlier version of this work were not re-derived
under the locked revision harness and the artifacts required to reproduce them were
not retained. No result in this paper depends on them; where a legacy value is cited, it
is as historical context for a limitation rather than as evidence. The deep \spsm{} and
\magspsm{} values reported here are the persisted lock-era values, which re-derive from
retained per-row prediction artifacts. A later revision reproduction gave a larger seed
dispersion; that reproduction survives only through summary-level artifacts and cannot be
independently re-derived from equivalently retained per-row prediction artifacts currently
preserved, no amendment authorised substituting it, and the cause of the difference remains
unresolved---nondeterministic GPU or environment behaviour is a hypothesis only. The locked protocol, its amendment chronology, the held-out data-access record, the environment-of-record summary, and the available artifact provenance are documented directly in the appendices of this paper. Retained analysis code will be made publicly available; some historical execution paths behind the persisted external-evaluation artifacts are not fully retained, so we do not claim that every reported artifact can be regenerated from released code. The persisted artifacts and their digests are unaffected by this limitation.

\appendix
\renewcommand{\thesection}{\alphalph{\value{section}}}

\section{Why the problem is hard: diagnostics and the magnitude construction}
\label{app:why-hard}
\label{sec:diagnostic}

\begin{figure}[t]
  \centering
  \includegraphics[width=0.95\linewidth]{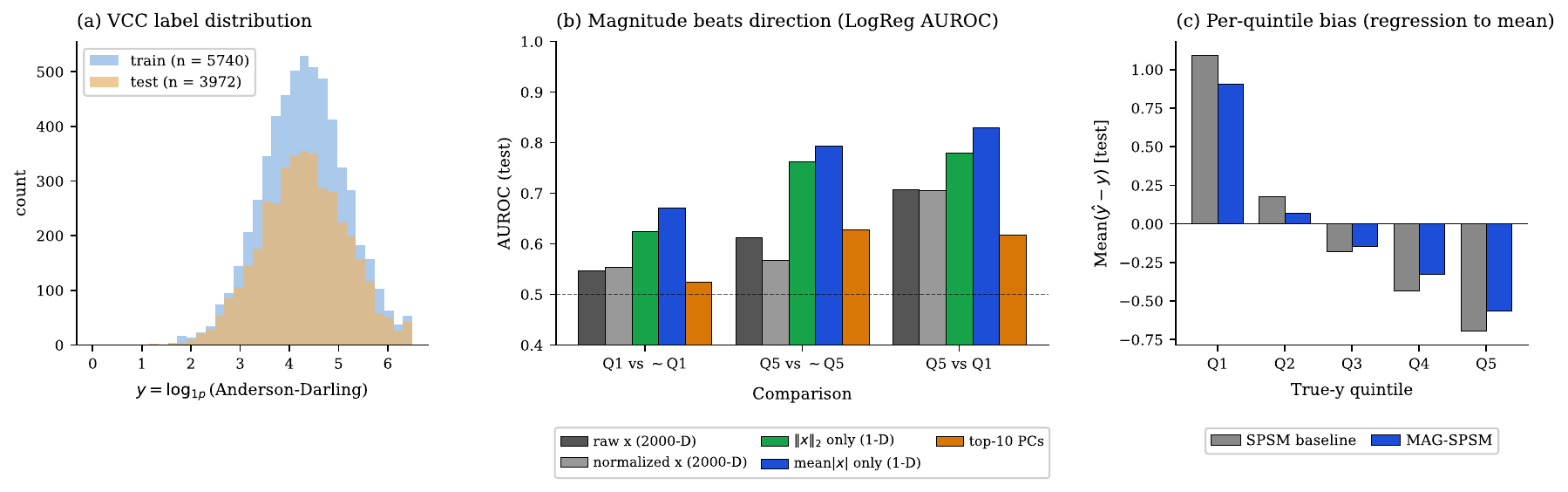}
  \caption{Diagnostic on VCC. \textbf{(a)}~Label distribution on the training and test
  splits. \textbf{(b)}~Separability of the highest from the lowest label quintile by
  logistic regression: one-dimensional magnitude summaries match or exceed the full
  $2{,}000$-dimensional expression vector and its leading principal components.
  \textbf{(c)}~Per-quintile bias: \spsm{} collapses predictions toward the marginal
  training mean; \magspsm{} reduces the bias most at the tails.}
  \label{fig:diagnostic}
\end{figure}

\paragraph{Baseline failure mode.}
The \spsm{} V4 baseline, a stable-prediction MLP encoder with a
within-context stability loss and a relation loss
(App.~\ref{app:spsm}), achieves test pooled $\rsq = +0.0824 \pm 0.0100$
across three seeds. Its predictions exhibit mean collapse: the prediction range is a small fraction of the
label range, and per-quintile bias is monotonic, over-predicting the lowest quintile and
under-predicting the highest (Fig.~\ref{fig:diagnostic}c). We say
``mean collapse'' rather than ``mode collapse'' \citep{wu2025perturbench}:
predictions concentrate at the marginal training mean rather than at one of
several modes of a multimodal output.

\paragraph{Five remedies that failed or stalled.}
Earlier exploratory work tried five plausible corrections under a matched protocol---an
inductive-bias bilevel gate, cross-cell-type pretraining followed by VCC fine-tuning, a
post-hoc tail correction, a tail-weighted MSE \citep{ren2022balancedmse}, and LDS-weighted
MSE \citep{yang2021delving}---and all failed or stalled. Those runs predate the locked
evaluation and their artifacts were not retained, so we use them as historical motivation
and draw no quantitative or mechanistic conclusion from them. What they motivated is the
step taken next: supplying response \emph{magnitude} to the predictor as an explicit
feature rather than through the loss.

\paragraph{The magnitude axis.}
A pre-specified diagnostic suite showed that one-dimensional magnitude summaries
separate low- from high-response rows about as well as the full expression vector,
while the baseline encoder collapses its predictions to a narrow window centered on
$\bar{y}_{\text{tr}}$ (Fig.~\ref{fig:diagnostic}). This motivates exposing magnitude
to the predictor explicitly.

\subsection{Magnitude features and a proof-of-concept encoder}
\label{sec:method}

\paragraph{Magnitude features.}
For $\xvec \in \Rb^{d}$ with components $x_1,\ldots,x_d$, define the
magnitude map $\magmap : \Rb^{d} \to \Rb^{4}$ by
\begin{equation}
\magmap(\xvec)
\;=\;
\Bigg(\,
\underbrace{\tfrac{1}{d}\!\sum_{j}\!|x_j|}_{m_1:\;\text{mean}|x|}
\,,\;
\underbrace{\Big(\sum_{j}\!x_j^{\,2}\Big)^{\!1/2}}_{m_2:\;\ell_2}
\,,\;
\underbrace{\max_{j}|x_j|}_{m_3:\;\ell_\infty}
\,,\;
\underbrace{\Big(\tfrac{1}{d}\!\sum_{j}(x_j-\bar{x})^{2}\Big)^{\!1/2}}_{m_4:\;\sigma(\xvec)}
\,\Bigg).
\label{eq:mag-features}
\end{equation}
We $z$-score $\magmap$ component-wise using \emph{training-set}
statistics to obtain $\magmaptilde(\xvec)$. Standardization removes any constant factor,
so specifying $m_2$ as $\|\xvec\|_2$ or as $\|\xvec\|_2/\sqrt{d}$ gives an identical
model input. All four statistics are
coordinate-permutation invariant, $m_1$--$m_3$ are sign-flip invariant, and
$m_4$ is invariant under translations along $\mathbf{1}$; these invariances
emphasize \emph{how much} a cell responded rather than \emph{which genes} did,
which motivates testing magnitude as a potentially more shift-robust summary across
cell types.

\paragraph{Three baselines built from \(\magmaptilde\).}
We use $\magmaptilde$ in three ways: (i) \maglinear{}, a four-feature
linear regression $\hat{y} = \wvec^{\top}\magmaptilde(\xvec) + b$;
(ii) classical methods on $\xvec$, $\xvec/\|\xvec\|_2$,
$\magmaptilde$, or $[\xvec;\,\magmaptilde]$
(App.~\ref{sec:strong-baselines}); and (iii) \magspsm{} (below) as a
proof of concept that the same signal can be exposed to a deep encoder.

\paragraph{\magspsm{} (proof of concept).}
\magspsm{} concatenates $\xvec$ with $\magmaptilde(\xvec)$ to form
$\xvec^{+} = [\xvec;\,\magmaptilde(\xvec)] \in \Rb^{2004}$ and feeds it
to an MLP encoder $f_\theta : \Rb^{2004} \to \Rb^{256}$ (three hidden
layers of width $256$, ReLU, LayerNorm, no architectural dropout)
followed by a linear head.
The objective is
\begin{equation}
\mathcal{L}_{\text{total}} \;=\; \mathcal{L}_{\text{MSE}}
\,+\, \lambda_{\text{stab}}\,\mathcal{L}_{\text{stab}}
\,+\, \lambda_{\text{rel}}\,\mathcal{L}_{\text{rel}},
\label{eq:total-loss}
\end{equation}
where $\mathcal{L}_{\text{stab}}$ and $\mathcal{L}_{\text{rel}}$ are
the \spsm{} stability and relation losses (App.~\ref{app:spsm};
$\lambda_{\text{stab}} = \lambda_{\text{rel}} = 0.5$, inherited
unchanged). Training schedule and optimizer are in
App.~\ref{app:spsm}; model selection is by validation $\rsq$.
\magspsm{} is \emph{not} the strongest model here---classical tree ensembles on
the same input outperform it (App.~\ref{sec:strong-baselines})---but is retained
as the simplest deep-architecture check that magnitude is recoverable when
supplied explicitly, and as the substrate for the per-row alignment controls
(App.~\ref{sec:causal}).

\subsection{Magnitude-anchored residual prediction}
\label{sec:mag_anchor}

\emph{MAG-Anchor} is a secondary deep architecture that splits the prediction into a
magnitude anchor and a gated residual,
$\hat{y} = g_\phi(\magmaptilde(\xvec)) + \alpha\, h_\theta([\xvec;\magmaptilde(\xvec)])$,
with $g_\phi$ initialized from MAG-LINEAR and $\alpha$ learned. It was not reproduced
under the locked revision harness and no value from it is carried into this revision;
its architecture and per-row alignment controls are in
App.~\ref{app:mag-anchor-full}.

\subsection{The role of per-row alignment}
\label{sec:causal}

The controls above rest on a simple
population-level argument: if an augmenting block $M^\star$ is
independent of $(X,Y)$, then
$\mathbb{E}[Y \mid X, M^\star] = \mathbb{E}[Y \mid X]$, so no
measurable predictor using $(X,M^\star)$ can improve over the Bayes
predictor on $X$ alone under squared loss. This yields two consequences we test
directly: (a)~replacing $\magmaptilde(\xvec_i)$ by four i.i.d.\
$\mathcal{N}(0, I_4)$ features makes the augmenting block independent
of $(\xvec_i, Y_i)$ by construction; (b)~a row-shuffled
augmentation preserves the marginal distribution of $\magmaptilde$ but
breaks row-specific information about $Y_i$. A systematic $\rsq$ gain over both
controls therefore supports the reading that the four scalars contribute through
per-row dependence on $\xvec_i$ rather than through added dimensionality alone
(App.~\ref{app:per-row-alignment}).


\section{The frozen foundation-model representation}
\label{app:fm-representation}

\paragraph{Model and pinned artifacts.}
The representation is produced by Geneformer-V2-104M
\citep{theodoris2023geneformer}, pretrained and used frozen with no fine-tuning:
twelve transformer layers, hidden width $768$, twelve attention heads, a
vocabulary of $20{,}275$ tokens, and a maximum sequence length of $4{,}096$. The
artifacts below were pinned before any result was produced; digests are the first
sixteen hexadecimal characters of the SHA-256 of each file. Full 64-character digests,
together with the environment-of-record summary, are reported in App.~\ref{app:environment-provenance}.

\begin{center}
\small
\begin{tabular}{ll}
\toprule
Artifact & Digest \\
\midrule
Geneformer repository commit & \texttt{ad8f66dfcda3ebbd} \\
V2-104M checkpoint, $417{,}571{,}156$ bytes & \texttt{fff5cba29ddd8792} \\
gc104M token dictionary & \texttt{67c445f4385127ad} \\
gc104M gene-median dictionary & \texttt{a51c53f6a771d645} \\
gc104M Ensembl mapping dictionary & \texttt{0819bcbd869cfa14} \\
\bottomrule
\end{tabular}
\end{center}

\paragraph{What ``frozen'' means here.}
The encoder is loaded in evaluation mode. No optimizer, backward pass, or
gradient-enabled encoder parameter appears anywhere in the analysis pipeline, and
re-embedding the same cells reproduces the stored features exactly, to a maximum
absolute difference of zero. The only trained components are the downstream heads
of App.~\ref{app:heads}.

\paragraph{Vocabulary coverage.}
The Geneformer token vocabulary covers $17{,}975$ of $18{,}080$ measured genes in VCC
($99.42\%$), $8{,}520$ of $8{,}749$ in RPE1 ($97.38\%$), and $8{,}210$ of $8{,}563$ in
K562-essential ($95.88\%$). Coverage is therefore high and comparable across the three
screens, so gross vocabulary loss is unlikely on its own to explain the negative external
transfer of Sec.~\ref{sec:transfer}. Other recorded cross-screen differences, set out
below, remain.

\paragraph{Input and tokenization.}
The input is the per-cell raw count vector. Genes are identified by Ensembl
accession and passed to the model's own tokenizer, which applies per-cell count
normalization and gene-wise median scaling and emits a rank-ordered token
sequence truncated to $4{,}096$ tokens. Duplicate, unmapped, and absent genes are
handled inside the tokenizer and its pinned mapping dictionary; we impose no
additional gene rule. Batching used a chunk size of $512$, which affects
throughput only.

\paragraph{Cell selection.}
Cells enter the embedding if they belong to an eligible construct or are
non-targeting. For VCC, constructs are additionally required to carry at least
ten cells in their batch and target-gene group. Beyond this we apply no additional explicit per-cell count, gene-count, or doublet
filter outside the tokenizer; any tokenizer-internal filtering is not separately
recorded. We state this as a limit of the record rather than as evidence that no
such filtering occurred.

\paragraph{Pooling and aggregation.}
Per cell we take the class-token activation at the second-to-last transformer
layer, giving $E_i \in \Rb^{768}$ in single precision. Padding is masked; no
other token masking is applied. Perturbation-level vectors are
\begin{equation}
  \zvec_g \;=\; \frac{1}{|g|}\sum_{i \in g} E_i \;-\; \mu_{\mathrm{NT}},
  \qquad \zvec_g \in \Rb^{768},
  \label{eq:z-def}
\end{equation}
where $E_i$ is the embedding of cell $i$ and $g$ ranges over the cells of one
construct. For the external screens the grouping key is the construct identifier;
for VCC, rows are joined on split, batch, and target gene.

\paragraph{The non-targeting reference.}
$\mu_{\mathrm{NT}}$ is the mean embedding of cells annotated non-targeting. Its
scope differs by screen, and the difference is recorded rather than incidental.
For VCC it is batch-local: one non-targeting mean per Flex batch, subtracted from
the constructs in that batch. For RPE1 it is a single globally pooled mean over
$11{,}485$ non-targeting cells, and for K562-essential a single globally pooled
mean over $10{,}691$, in both cases with no batch matching and no subsampling.

\paragraph{What is identical across screens, and what the record shows is not.}
The extraction configuration was verified identical across VCC, RPE1, and
K562-essential in checkpoint, embedding mode, layer, numerical precision,
sequence length, chunking, subtraction form, and output type. Four things are
\emph{not} identical, and we state them because each is a possible alternative
explanation for the external results of Sec.~\ref{sec:transfer}.

\begin{enumerate}[itemsep=2pt,topsep=2pt,leftmargin=*]
\item \emph{Scope of the non-targeting reference}---batch-local for VCC, globally
pooled for both external screens.
\item \emph{Size of the non-targeting pool}---$11{,}485$ cells for RPE1 against
$10{,}691$ for K562-essential.
\item \emph{Provenance of the embeddings}---those for VCC and for $1{,}889$ of
the $1{,}932$ RPE1 constructs were read from a precomputed store generated with
the same pinned checkpoint and the same extraction configuration; the remaining
$43$ RPE1 constructs and all K562-essential constructs were embedded afresh. The value $43$
appears in three distinct contexts in this manuscript: raw VCC-training-target overlap, RPE1
constructs newly embedded during top-up, and RPE1 constructs for which the batch-matched control
is computable. These refer to different construct sets and should not be conflated.
Under the recorded extraction configuration, deterministic re-embedding showed
numerical agreement, so cached and freshly computed embeddings were treated as
equivalent for aggregation.
\item \emph{RPE1 cell sets}---the RPE1 embedding set and the RPE1 cell-count
accumulator used elsewhere in the paper are distinct objects produced by
different quality filters, with minima of $27$ and $25$ cells per construct and
non-targeting pools of $11{,}485$ and $11{,}007$ cells respectively.
\end{enumerate}

\noindent
We did not test whether the asymmetry in the non-targeting reference contributes
to the negative external transfer reported in Sec.~\ref{sec:transfer}. The frozen
protocol admits no post-unblinding re-analysis, so this remains a live
alternative explanation and we flag it as such rather than dismissing it.

\paragraph{Scaling after construction.}
Each feature block carries its own standardizer---centering and scaling to unit
variance---fit on that block's VCC training rows alone and applied with frozen
statistics to validation, test, RPE1, and K562-essential. The representation and
each control carry separate standardizers. No $\ell_2$ normalization, whitening,
principal-component projection, or clipping is applied to $\zvec$; the clipping
referenced elsewhere in this paper applies only to the expression and magnitude
blocks.

\paragraph{Determinism and seeds.}
No random seed governs the construction of $\zvec$. Seeds enter only downstream:
head training, the secondary matched-subsampling sensitivity analysis, and the
bootstrap intervals of App.~\ref{app:uncertainty}.

\paragraph{A limit on provenance.}
The persisted result records do not carry a content digest binding a specific
stored copy of $\zvec$ to the reported external correlations; the link between
the feature arrays and the persisted predictions is recorded in prose rather than
in a manifest. What \emph{is} pinned is the complete generating specification---
checkpoint, repository commit, tokenizer and vocabulary digests, layer, pooling,
aggregation, and scaling---together with the observed determinism of the embedding
step. That specification supports deterministic regeneration under the recorded
pipeline, but we cannot demonstrate that a regenerated array is bitwise identical to
the exact feature array behind the reported correlations. We state this rather than
implying an archival guarantee we cannot support.

\section{Environment of record and full artifact provenance}
\label{app:environment-provenance}

This appendix records the environment information and full artifact identifiers needed to interpret the frozen-representation results. We distinguish values written by the governed run (\textsf{RUN\_RECORD}) from values read from the machine during the later revision (\textsf{CURRENT\_MACHINE}); the latter are not treated as evidence about the original embedding run unless a run record independently supports them.

\paragraph{Foundation-model embedding environment (RUN\_RECORD).}
The persisted smoke-gate manifest and extraction logs record Python 3.10.20, PyTorch 2.3.1+cu121, and \texttt{transformers} 4.49.0 for the embedding run. A forward-looking note in the pre-result specification had estimated \texttt{transformers} as approximately 4.44--4.46; the run-written value 4.49.0 is the environment that actually produced the embeddings. No portable \texttt{environment.yml} or \texttt{requirements.txt} survives for that run, which is a reproducibility limitation.

\paragraph{Revision analysis environment (CURRENT\_MACHINE).}
The later classical-head, bootstrap, and tabulation work was carried out in a separate project environment. The revision machine recorded Python 3.13.12, PyTorch 2.11.0+cu130, NumPy 2.4.4, pandas 2.3.3, SciPy 1.17.1, scikit-learn 1.8.0, anndata 0.12.11, scanpy 1.12.1, and h5py 3.16.0. These package versions describe the revision machine, not the original embedding run. The machine map recorded two NVIDIA TITAN RTX GPUs (24 GB each), driver 580.126.20, and CUDA driver 13.0; the embedding run itself used a cu121 PyTorch build, and its GPU model is not independently written in the run manifest.

\paragraph{Pinned upstream artifacts (RUN\_RECORD).}
The following full identifiers were pinned before results were produced. For readability, SHA-256 strings are split into four 16-character groups; concatenating the groups yields the exact digest.

\begin{center}
\scriptsize
\begin{tabular}{@{}p{0.25\linewidth}p{0.67\linewidth}@{}}
\toprule
Artifact & Full identifier \\
\midrule
Geneformer repository commit & \texttt{ad8f66dfcda3ebbd148d916c01f31339c5b95a15} \\
V2-104M checkpoint & \texttt{fff5cba29ddd8792 991fa77b4872246f be548a178cebda37 75cdc72b67780e7f} \\
Token dictionary & \texttt{67c445f4385127ad fc48dcc072320cd6 5d6822829bf27dd3 8070e6e787bc597f} \\
Gene-median dictionary & \texttt{a51c53f6a771d645 08dfaf61529df70e 394c53bd20856926 117ae5d641a24bf5} \\
Ensembl mapping dictionary & \texttt{0819bcbd869cfa14 279449b037eb9ed1 d09a91310e77bd1a 19d927465030e95c} \\
\bottomrule
\end{tabular}
\end{center}

\paragraph{Harmonized endpoint archives.}
The archived per-construct harmonized outcomes used by the external analyses have SHA-256 digests \texttt{353c04d3ac8e3380 a3addf3abc1953d3 0799423f7cd46700 6adc36800e26a126} for RPE1 (1,932 rows) and \texttt{051269eeeae93c75 c84ca1fc050529a0 96ebd15e1990cb14 bb24669c1b9672ff} for K562-essential (1,903 rows). These hashes identify the persisted outcome vectors used for the primary external transfer analysis and the non-overlap sensitivity.

\paragraph{Existing-asset licenses and access terms.}
We record the terms we can verify rather than infer a license where none is stated. The Geneformer-V2-104M model repository identifies the model as Apache-2.0 licensed. The Replogle processed Perturb-seq release used here (Figshare+ DOI 10.25452/figshare.plus.20029387) identifies the dataset license as CC BY 4.0. The \texttt{scikit-learn} software used for the classical baselines is distributed under the BSD 3-Clause license. For the 2025 Virtual Cell Challenge H1 hESC benchmark, the public Arc challenge and Virtual Cell Atlas pages document access to the released train/validation/test data and the applicable challenge/access rules, but we did not locate a standalone dataset-license identifier on those public pages; we therefore do not assign one here. None of these upstream assets is redistributed with this paper.

\section{Control definitions}
\label{app:controls}

All five controls were specified in the pre-result lock. Feature-generation seeds
are disjoint from head-training seeds, and every control is generated from array
shapes alone: no label, no validation statistic, and no test statistic enters any
control. Below, the split index runs over training, validation, test, RPE1, and
K562-essential in that order, and each control's generator is seeded at its base
value plus that index.

\paragraph{$C_1$: the capacity floor (gating).}
A matrix in $\Rb^{n \times 768}$ with independent standard Gaussian entries drawn
per row in single precision, base seed $20270101$, one deterministic draw per
split, generated independently within each split from the row count alone.
External splits receive their own draws; nothing is transferred from VCC. The
expected row norm is $\sqrt{768} \approx 27.71$. Exactly one realization exists
per split, so reported values are not averaged over several Gaussian draws and the floor
is conditioned on that one pre-specified draw. Drawing more would have required choosing,
after the fact, how many to draw and how to combine them, which the specification does not
fix; we note the dependence rather than resolve it after unblinding.
``Capacity-matched'' here means matched to $\zvec$ in input dimension, and
sharing its preprocessing, candidate-head pool, selection rule, seeds, and
training budget. It is a zero-information floor, and dimension matching is the
criterion the specification enforces; we claim no stronger notion of equal
effective capacity.

\paragraph{$C_2$: compressed expression benchmark (secondary, non-gating).}
A random projection of the $2{,}000$-dimensional expression vector to $768$
dimensions with independent Gaussian entries of variance $1/768$, drawn once with
seed $20270102$ and applied unchanged to every split.

\paragraph{$C_3$: alignment control.}
The representation of Eq.~\ref{eq:z-def} with its rows permuted. The permutation is drawn per
split from base seed $20270103$ and applied to that split's rows; the
\emph{same} permutation is used at training and at evaluation within a split, and
different permutations are used across splits. The permutation is unconstrained,
and the standardizer fit on the unpermuted representation is reused, so the
marginal distribution of $\zvec$ is preserved exactly and only the row-to-label
correspondence is destroyed.

\paragraph{$C_4$: alignment control on the magnitude block.}
The concatenation $[\,\zvec\,;\,\text{row-permuted } \magmaptilde\,] \in
\Rb^{n \times 772}$, with the magnitude rows permuted per split from base seed
$20270104$.

\paragraph{$C_5$: dimension-increment control.}
The representation concatenated with four independent standard Gaussian scalars,
giving a matrix in $\Rb^{n \times 772}$, drawn per row per split from base seed
$20270105$. Together, $C_4$ and $C_5$ separate a real contribution of the
magnitude block from the mere addition of four columns.

\paragraph{Pre-registration.}
All five controls appear in the original pre-result specification. The recorded
amendments govern external baselines, the external head, cell-count-matching
seeds, and secondary analyses; the remainder are editorial or record-keeping, and
none adds, removes, or reseeds a control. The
specification forbids introducing a control after primary results and forbids
changing a control seed because a result looks unusual.

\section{Head families, selection, and freezing}
\label{app:heads}

\paragraph{Candidate pool.}
Four head families with fixed configurations---not a grid, and no hyperparameter
search of any kind: ridge regression with unit regularisation; a random forest of
$200$ trees with maximum depth $20$ and at least two samples per leaf; histogram
gradient boosting with $500$ boosting iterations, maximum depth $6$, learning
rate $0.05$, and early stopping enabled; and a $256$-unit multilayer perceptron.
Each is a single-output regressor on $y$ and differs across arms only in input
width, $768$ or $772$.

\paragraph{Budget.}
Seven arms, four families, and three training seeds give $84$ fits, identical for
every arm.

\paragraph{Selection rule.}
Per arm, the argmax of the seed-mean VCC \emph{validation} $\rsq$. No test metric
influences selection. The rule is identical for $\zvec$, for
$[\zvec;\magmaptilde]$, and for every control.

\paragraph{Realised selections and their asymmetries.}
Under this single rule the arms resolved to different families: a random forest
for $\zvec$, $[\zvec;\magmaptilde]$, $C_2$, $C_4$, and $C_5$, and histogram
gradient boosting for $C_1$ and $C_3$. The multilayer perceptron diverged on
every arm except $C_2$, so the effective candidate pool was three families for
six arms and four for $C_2$. We report these asymmetries because they are the
natural first objection to the primary contrast: the procedure was symmetric, the
outcomes were not.

\paragraph{Freezing.}
The selection was written to disk before any VCC test label was scored. The
ordering is established by the within-process write order and by the execution
log, in which the freeze record precedes every test entry, rather than by
filesystem timestamps. Test \emph{predictions} were materialized during training,
before the freeze; only the scoring of predictions against test labels occurs
afterwards, and selection consumed validation predictions alone.

\paragraph{Robustness of the selection.}
The executed selection used an unguarded sort, so the ranking of divergent
entries was not explicitly specified. In this run the divergent multilayer
perceptron entries ranked below all finite maxima, and the selected family is the
argmax over finite validation scores for every one of the seven arms. We record
the fragility because it was real, and the outcome because it is verifiable from
the persisted validation table.

\section{Pre-registered hypotheses}
\label{app:hypotheses}

Four hypotheses were fixed in the pre-result specification, together with their
gating controls, their inference procedures, and their pass criteria. We
reproduce each criterion and report its outcome.

\paragraph{H-FM1: in-distribution information.}
Criterion: the frozen representation is informative if and only if the paired
bootstrap interval on $\Delta\rsq = \rsq(\text{FM-only}) - \rsq(C_1)$ excludes
zero on the positive side, with $C_1$ alone as the gate. Inference: row-level
paired bootstrap, $B = 10{,}000$, seed $42$, $95\%$ percentile interval
(App.~\ref{app:uncertainty}). \textbf{Outcome: passed}, with
$\Delta\rsq = +0.1645$ and interval $[+0.1375, +0.1920]$. Reported alongside and
non-gating, as the specification requires: $\Delta\rsq(\zvec-C_2) = +0.1107$,
interval $[+0.0906, +0.1317]$.

\paragraph{H-FM2: zero-shot transfer.}
Criterion: interpretable \emph{only if} the $C_1$ gate of H-FM1 passed. The
pre-registered expectation was that the frozen representation would behave like
the expression-only group, with an external Spearman correlation negative or
unresolved on \emph{both} screens. The specification further states that if the
gate fails, a near-zero external correlation must not be reported as a transfer
finding; the admissible statement in that case would have been that usable
in-distribution information had not been established. \textbf{Outcome: the gate
passed, so the external result is interpretable}, and the pre-registered
expectation was met, with $\rho = -0.139$ on RPE1 and $-0.267$ on
K562-essential, negative and resolved on both screens. The expected direction was
recorded before the first documented external-outcome access.

\paragraph{H-FM3: magnitude increment.}
Criterion: compare the magnitude-augmented representation against the frozen
representation on both screens, with the paired difference in Spearman
correlation as the primary quantity; a resolved positive increment is required on
\emph{both} screens, mirroring the increment previously observed from \spsm{} to
\magspsm{}, and the increment must survive $C_4$ and $C_5$. If a row-permuted
magnitude block or four Gaussian scalars reproduce the gain, no claim of per-row
magnitude alignment may be made, and no single-screen success may be claimed.
Inference: paired target-gene cluster bootstrap, $B = 10{,}000$, $95\%$
percentile interval. \textbf{Outcome: the criterion was met on both screens},
with $\Delta\rho = +0.032$ $[+0.008, +0.056]$ on RPE1 and $+0.143$
$[+0.111, +0.175]$ on K562-essential, both resolved positive and both surviving
$C_4$ and $C_5$. The RPE1 increment falls below, and the K562-essential increment
above, the mirroring values quoted in the pre-registered expectation; the
criterion was a resolved positive increment on both screens rather than a
magnitude threshold, and we do not reinterpret it as one.

\paragraph{Pre-specified non-overlap sensitivity.}
A small fraction of externally evaluated constructs target genes that also occur among
the $150$ raw VCC training targets: $43$ of $1{,}932$ on RPE1, spanning $42$ genes, and
$33$ of $1{,}903$ on K562-essential, spanning $32$ genes. The exclusion universe is the
raw training targets, not the $140$ that survive the eligibility filter of
Sec.~\ref{sec:setup}, and the two are not interchangeable here: the eligible set would
have removed $37$ constructs on RPE1 and $29$ on K562-essential rather than $43$ and $33$;
these counterfactual counts were reconstructed under a frozen metadata-only procedure
from the pre-specified set definitions, after first reproducing the historical $43$ and
$33$ counts (App.~\ref{app:chronology-evidence}).
Excluding on the raw set is the more conservative of the two rules, simply because the raw
set contains the eligible one and therefore removes strictly more. The additional six
constructs on RPE1 and four on K562-essential carry no exposure risk in any case: a target
gene that the cell-count floor removed from training contributes no training row, so the
model never saw it. Occurring among the raw $150$ is therefore not the same as having been
seen during training.
The specification fixes the primary on all eligible
constructs and pre-registers a secondary that repeats the external comparison with these
constructs removed, so that transfer can be read on target genes the model has never seen.
This secondary was not executed at the time the primary was persisted. It was run
afterwards, after external unblinding, and that ordering is recorded in the amendment chain.
The specification fixes no inference procedure for it, so we inherited the primary's: a
paired target-gene cluster bootstrap with $B = 10{,}000$, percentile intervals, and the
primary's seeds. Before subsetting, re-scoring the full sets from the stored predictions
reproduced every reported correlation, interval, and contrast exactly, so any difference
below is attributable to the exclusion rather than to the re-scoring.

On the retained sets---$1{,}889$ constructs for RPE1 and $1{,}870$ for K562-essential---no
verdict changes. The frozen representation remains negative on both screens under the inherited interval criterion
($\rho = -0.147$ and $-0.273$) and remains below its capacity floor with the paired intervals excluding zero; $C_1$ and
$C_3$ remain compatible with null; and the magnitude increment remains resolved positive
($\Delta\rho = +0.033$ and $+0.135$). No quantity moves by more than $0.014$, and the
external correlations shift slightly away from zero rather than toward it. This outcome is
inconsistent with target overlap being the explanation for the negative transfer. The shifts
are small relative to the intervals, and we do not read a direction into their magnitude.

\paragraph{H-FM4: magnitude recoverability (mechanism diagnostic).}
Criterion: four separate single-output ridge probes, one per magnitude scalar
rather than a shared multi-output fit, trained on the VCC training split with
regularisation selected on validation. Each probe is read against matched
references from the same probe family---a ceiling given by predicting the scalar
from the expression vector, and two floors given by a random projection of the
expression vector and by the row-permuted representation. No absolute threshold
was set: all four components were to be reported by their position between the
matched floors and the ceiling, without summarizing to a single number and
without cherry-picking. \textbf{Outcome: evaluated on train/validation only.}
The VCC test split was closed before the budgeted test-level evaluation could be
run; the diagnostic was later executed under the frozen design on the training
and validation splits alone, as a supporting diagnostic. The test-level statement
therefore remains unresolved, and the mechanism it was designed to probe---whether
$\zvec$ carries the magnitude information that $\magmaptilde$ supplies---is not
settled by train/validation evidence. We report this rather than omitting it,
because a pre-registered analysis and the restricted form in which it was
eventually run are both part of the record.

\subsection{Locked external evaluation, in full}
\label{app:locked-external}

Table~\ref{tab:fm_external_primary} gives the current locked external evaluation for every
pre-registered arm and contrast on both screens. It is the evaluation Sec.~\ref{sec:transfer}
reports; the legacy classical-transfer table elsewhere in this appendix is a separate,
historical object and is not this evaluation.

\begin{table}[t]
\centering
\caption[Zero-shot external transfer of the frozen representation]{Primary zero-shot external evaluation:
frozen VCC heads applied unchanged (no target-side calibration) to RPE1 and K562-essential on the
construction-matched $\xvec_{\mathrm{cell}}$ basis. Entries are construct-level Spearman $\rho$ against
the harmonized target-gene endpoint with $95\%$ paired target-gene cluster-bootstrap intervals
($B=10{,}000$; per-screen seeds RPE1 $20260901$, K562 $20260902$). $\zvec$ lands \emph{below} its
capacity-matched control C1 on both screens.}
\label{tab:fm_external_primary}
\small
\setlength{\tabcolsep}{4pt}
\begin{tabular}{lcc}
\toprule
Arm & RPE1 $\rho$ [95\% CI] & K562-essential $\rho$ [95\% CI] \\
\midrule
$\zvec$ (FM-only) & $-0.139$ $[-0.185, -0.092]$ & $-0.267$ $[-0.310, -0.224]$ \\
$[\zvec;\magmaptilde]$ & $-0.107$ $[-0.152, -0.062]$ & $-0.124$ $[-0.170, -0.080]$ \\
C1 (Gaussian) & $-0.034$ $[-0.078, +0.012]$ & $+0.032$ $[-0.012, +0.077]$ \\
C2 (proj. $\xvec$) & $-0.156$ $[-0.202, -0.109]$ & $-0.264$ $[-0.306, -0.221]$ \\
C3 (shuffled $\zvec$) & $+0.044$ $[-0.001, +0.088]$ & $-0.004$ $[-0.049, +0.041]$ \\
C4 & $-0.138$ $[-0.184, -0.091]$ & $-0.266$ $[-0.309, -0.223]$ \\
C5 & $-0.139$ $[-0.184, -0.091]$ & $-0.269$ $[-0.312, -0.226]$ \\
\midrule
\multicolumn{3}{l}{\emph{Pre-registered paired contrasts ($\Delta\rho$, 95\% CI)}} \\
$\zvec-$C1 & $-0.106$ $[-0.168, -0.044]$ & $-0.299$ $[-0.360, -0.238]$ \\
$[\zvec;\magmaptilde]-\zvec$ (H-FM3) & $+0.032$ $[+0.008, +0.056]$ & $+0.143$ $[+0.111, +0.175]$ \\
$[\zvec;\magmaptilde]-$C4 & $+0.031$ $[+0.007, +0.056]$ & $+0.142$ $[+0.111, +0.174]$ \\
$[\zvec;\magmaptilde]-$C5 & $+0.031$ $[+0.008, +0.056]$ & $+0.144$ $[+0.113, +0.177]$ \\
\bottomrule
\end{tabular}
\end{table}

\subsection{Pre-specified secondary external analyses}
\label{app:secondaries}

The locked protocol contained several pre-specified external secondary analyses. The two
reporting obligations omitted from the previous manuscript version were A1's
$\xvec_{\mathrm{release}}$ feature-basis comparisons and A2's ordinary linear head on
$[\zvec;\magmaptilde]$; Table~\ref{tab:fm_secondary} also reports the two pre-specified A4
secondaries, MAX-LINEAR and MAG+COUNT. All were fixed before external evaluation and executed
only after the frozen primary had been persisted, following the recorded execution ordering.
All remain secondary: none replaces the frozen primary result, and disagreements are reported
rather than resolved by post-outcome selection.

Their scope is bounded by the amendments that fixed them. The A2 linear head applies only to
arms containing $\magmaptilde$, not to $\zvec$-only arms; it therefore re-evaluates no
hypothesis, in particular not H-FM3, and leaves the $\zvec$-only external results of
Sec.~\ref{sec:transfer} untouched. The A1 rows repeat the corresponding external arm and control
definitions on the released $\xvec_{\mathrm{release}}$ feature basis rather than the
frozen-primary $\xvec_{\mathrm{cell}}$ basis; the $\xvec_{\mathrm{cell}}$ counterparts are in
Table~\ref{tab:fm_external_primary}, so the two bases can be compared arm by arm. What these
establish is that the magnitude-augmented arm is sensitive to the pre-specified choice of head
and of feature basis: the linear head is positive on RPE1 with its interval excluding zero and inconclusive on
K562-essential, while the $\xvec_{\mathrm{release}}$ basis weakens the negative RPE1 association
and is inconclusive on K562-essential. The persisted paired comparison between the two bases is
in the table caption.

\paragraph{MAX-LINEAR.}
MAX-LINEAR is the pre-specified one-feature zero-shot linear predictor fit on VCC train using
$\max_j |x_j|$ alone. It is distinct from the count-adjusted partial-Spearman diagnostic of
$\max_j |x_j|$ in Sec.~\ref{sec:max-response}, which partials out $\log(1 + n_{\text{cells}})$;
the two are different estimands and should not be read as one quantity. MAX-LINEAR is positive on
both external screens, whereas the frozen $\zvec$-only predictor is negative. Because MAX-LINEAR
uses a different one-dimensional magnitude signal and readout, this contrast shows that external
transfer behaviour depends on the predictive signal and estimator being evaluated; it does not
identify the mechanism of the $\zvec$-only failure, and it is not promoted.

\paragraph{MAG+COUNT, and a limitation recorded in advance.}
MAG+COUNT appends $\log(1 + n_{\text{cells}})$ to the magnitude block. The external count ranges
extend far beyond the VCC-training range (VCC $10$--$121$; RPE1 $25$--$3{,}451$; K562-essential
$27$--$1{,}996$), so the frozen external linear head extrapolates on this feature. This limitation
was specified before execution and is reported here alongside the result; it is not an explanation
inferred from the observed outcome.

\paragraph{A metadata caution.}
The persisted secondary artifact records a cell-set field,
\texttt{RPE1\_cellset\_for\_counts}, identifying the checkpoint population behind its
count-derived quantities. That field describes count provenance only. It is not a qualifier on the
correlations reported here, which are computed over the persisted evaluated construct sets, and it
is a different population from the persisted-embedding cell set used by the matched sensitivity of
App.~\ref{app:matched-sensitivity}. No separately persisted paired contrast between any of these
secondaries and the $\zvec$-only arm is reported here.

\begin{table}[t]
\centering
\caption[Pre-specified secondary external analyses]{Pre-specified secondary external analyses
(construct-level Spearman $\rho$ with $95\%$ paired target-gene cluster-bootstrap intervals,
$B = 10{,}000$). Every row is a \textsc{secondary} analysis; none is promoted, and none replaces a
frozen primary result. The three A1 rows repeat the corresponding external arm and control
definitions on the released $\xvec_{\mathrm{release}}$ feature basis; their frozen-primary
$\xvec_{\mathrm{cell}}$ counterparts are in Table~\ref{tab:fm_external_primary}. For
$[\zvec;\magmaptilde]$ the persisted $\xvec_{\mathrm{cell}}$ minus $\xvec_{\mathrm{release}}$
paired contrasts are $-0.056$ $[-0.084, -0.028]$ on RPE1 and $-0.131$ $[-0.179, -0.084]$ on
K562-essential; the disagreement is reported without selecting either basis post hoc.
MAX-LINEAR is the pre-specified one-feature zero-shot linear predictor fit on VCC train using
$\max_j |x_j|$ alone; it is distinct from the count-adjusted partial-Spearman diagnostic of
$\max_j |x_j|$ in Sec.~\ref{sec:max-response}, which partials out $\log(1 + n_{\text{cells}})$.
The intervals shown for MAX-LINEAR are the persisted intervals for this zero-shot predictive
secondary; they are not intervals for that separate diagnostic, for which no interval procedure
was pre-specified or attached.
MAG+COUNT appends $\log(1 + n_{\text{cells}})$ to the magnitude block; the external count ranges
extend far beyond the VCC-training range (VCC $10$--$121$; RPE1 $25$--$3{,}451$; K562-essential
$27$--$1{,}996$), so the frozen external linear head extrapolates on this feature. That limitation
was specified before execution and is recorded alongside the number rather than inferred from it.}
\label{tab:fm_secondary}
\small
\setlength{\tabcolsep}{5pt}
\begin{tabular}{lcc}
\toprule
Secondary analysis & RPE1 $\rho$ [95\% CI] & K562-essential $\rho$ [95\% CI] \\
\midrule
A4: MAX-LINEAR & $+0.232$ $[+0.185, +0.278]$ & $+0.354$ $[+0.310, +0.395]$ \\
A4: MAG+COUNT & $-0.027$ $[-0.072, +0.017]$ & $+0.033$ $[-0.013, +0.077]$ \\
A2: linear head on $[\zvec;\magmaptilde]$ & $+0.097$ $[+0.054, +0.140]$ & $+0.020$ $[-0.025, +0.064]$ \\
\midrule
A1: $[\zvec;\magmaptilde]$ on $\xvec_{\mathrm{release}}$ & $-0.052$ $[-0.097, -0.006]$ & $+0.007$ $[-0.041, +0.054]$ \\
A1: $C_2$ on $\xvec_{\mathrm{release}}$ & $-0.175$ $[-0.221, -0.128]$ & $-0.274$ $[-0.317, -0.231]$ \\
A1: $C_4$ on $\xvec_{\mathrm{release}}$ & $-0.139$ $[-0.184, -0.091]$ & $-0.267$ $[-0.310, -0.224]$ \\
\bottomrule
\end{tabular}
\end{table}

\subsection{Position in the pre-specified interpretation table}

The specification fixed a two-way interpretation table before any number existed, crossing
whether $\zvec$ transfers well or poorly with whether $\magmaptilde$ is recoverable from
$\zvec$. Under that table the observed result occupies the ``$\zvec$ transfers poorly'' row.
The recoverability column remains undetermined, because H-FM4 was evaluated on
train/validation only and was never run against the held-out external outcomes; accordingly,
neither test-level column is assigned. No external H-FM4 evaluation exists.

\section{Sampling-depth diagnostics}
\label{app:depth-diagnostics}

\subsection{A pre-specified feature-side sensitivity before documented
external-outcome access}
\label{app:matched-sensitivity}

\paragraph{Why this exists, and when it was fixed.}
The harmonized external endpoint is built from fixed-size subsamples, but the magnitude
and representation blocks are means over all of a construct's cells, so cells per construct
differ across screens by construction. A recorded amendment fixed a sensitivity for exactly
this asymmetry on 2026-08-13, before any of its own measurements existed. The surviving local
record places this sensitivity before the first documented external-outcome access.

\paragraph{Design.}
Two matched cell counts were fixed mechanically from the observed per-construct minima,
with no discretionary choice: $\nmatch = 10$, the feature path's existing floor, at which
every construct in every screen qualifies; and $\nmatch = 27$, the maximum of the three
screens' minima ($10$, $25$, $27$) in the table available at specification time. For RPE1
that minimum of $25$ was measured on the checkpoint $\xvec_{\mathrm{cell}}$ accumulator; the
executed construction-matched sensitivity instead operated on the persisted-embedding cell
set, whose RPE1 minimum is $27$. For each construct and each draw, $\nmatch$ perturbed
cells are sampled once and \emph{both} blocks are rebuilt from that same cell set, so
construction parallelism is preserved. The non-targeting reference is never subsampled:
resampling it would perturb a second noise source that enters every construct in a screen.
Three sampling seeds were pre-registered ($20270201$, $20270202$, $20270203$). No embedding
was recomputed---the representation is re-aggregated from the persisted per-cell
embeddings.

This is a different intervention from the endpoint's fixed-size subsampling of
App.~\ref{app:harmonized-endpoint}. The two share the number $27$ by coincidence of
derivation---the endpoint's is a VCC median, this one a maximum over screen minima---and
they act on opposite sides of the pipeline. They should not be conflated.

\paragraph{Retention is part of the intervention, and was declared in advance.}
At $\nmatch = 10$ every construct is retained in all three screens ($5{,}740$ VCC,
$1{,}932$ RPE1, $1{,}903$ K562-essential). At $\nmatch = 27$, VCC retains $2{,}843$ of
$5{,}740$ and drops $2{,}897$, while RPE1 retains $1{,}932$ of $1{,}932$ (none dropped) and
K562-essential $1{,}903$ of $1{,}903$. The drop the amendment anticipated for RPE1's
$25$--$26$-cell constructs therefore did not occur on the executed population. The
amendment stated before the measurement that VCC's median is $26$, that roughly half of it
would therefore drop, and that the survivors would be its high-count constructs: the
retained VCC median rises from $26$ (IQR $[18,40]$) to $40$ (IQR $[32,50]$), while RPE1 and
K562-essential are unselected at $76$ (IQR $[50,118]$) and $118$ (IQR $[79,174]$). The RPE1
summary here---median $76$, IQR $[50,118]$---is that of the persisted-embedding cell set this
sensitivity used; the $73$ with IQR $[48,113]$ reported for the sampling-depth accumulator
(Fig.~\ref{fig:sampling-depth}C) describes the distinct checkpoint
$\xvec_{\mathrm{cell}}$ population over the same $1{,}932$ constructs. Only
VCC's retained set is a biased high-count sample. This is a declared property of the
intervention rather than a discovery about it, and both counts are reported precisely
because they trade off in opposite directions: $n = 10$ carries no retention-induced
selection but leaves more estimation noise, while $n = 27$ leaves less noise but selects on
cell count.

\paragraph{Result: the ordering changes with the matched count.}
Over all cells the cross-screen ordering of median magnitude is
K562-essential $<$ VCC $<$ RPE1. At \emph{both} matched counts it becomes
VCC $<$ K562-essential $<$ RPE1: RPE1 stays largest while VCC and K562-essential exchange
places. The change is stable across all three seeds with non-overlapping intervals, and the
representation block shows the same reordering as the magnitude block. Under the
interpretation fixed before these numbers existed, an ordering that changes at either count
makes the cross-screen magnitude comparison sampling-depth dependent and obliges us to
state that the magnitude features, unlike the endpoint, are not size-matched across
screens. We state it.

\paragraph{What this does not show.}
It does not establish a causal role for sampling depth in the external transfer failure, it
does not show that matching features repairs transfer, and it licenses no claim of
robustness to sampling depth. A matched-feature external \emph{evaluation} was deliberately
placed outside the amendment's scope; it would have required its own amendment committed
before the external labels were opened, and it was never run. What is established is
narrower and wholly descriptive: a comparison this paper makes across screens moves when
feature-side sampling depth is equalised, so that comparison should be read as
depth-dependent.

\subsection{Two later diagnostics, recorded and executed after unblinding}
\label{app:post-unblinding}

Both diagnostics below were specified and executed on 2026-08-29, after external
unblinding. Each was locked before its own result existed, but neither belongs to the
original pre-result specification, neither gates or revises any hypothesis, and neither
touches the VCC test split or any external outcome. Both are classified
\textsc{secondary}\slash\textsc{exploratory} under the specification's multiplicity rule,
which reserves the primary inferential family for the four pre-registered hypotheses: neither
is promoted into that family, and neither enters any composite score. We report them because
they bear on limitations stated elsewhere in this paper.

\paragraph{A linear depth-recovery probe on the representation.}
Target: $\log(1 + n_{\text{cells}})$ over a construct's perturbed cells, truncated at the
ten-cell eligibility floor and so supported on $[10, 121]$ cells. Data: VCC training and
validation rows only ($5{,}740$ and $2{,}052$ rows, the latter drawn from $49$ target genes). Predictor: the frozen representation $\zvec$ alone. Estimator:
ridge regression over the pre-specified penalty grid
$\{10^{-3}, 10^{-2}, 10^{-1}, 1, 10, 10^{2}, 10^{3}\}$, selected on validation $\rsq$.
Validation $\rsq$ rose monotonically across the grid and stayed negative throughout, from
$-0.140$ at $10^{-3}$ to $-0.015$ at $10^{3}$; the selected penalty was the grid's upper
endpoint, $10^{3}$.

Both clauses of the result travel together: this probe did not linearly recover
sampling-depth-associated information from the frozen representation on VCC
train/validation under the pre-specified estimator and grid, \emph{and} the regularisation
optimum was not bracketed above within that grid, so non-recovery is not established even
for this estimator family. It is in particular not evidence that $\zvec$ fails to encode
sampling depth, that depth information is absent, or that the depth explanation considered
in Sec.~\ref{sec:discussion} has been ruled out. The grid was frozen in advance and was not
widened after the boundary selection was observed.

Three bounds travel with that number. No interval, bootstrap, $p$-value or success threshold
was specified for this probe and none is attached here, so the reported $\rsq$ is a single
deterministic value and the validation figure rests on $49$ gene clusters rather than
$2{,}052$ independent rows. The probe asks whether $\zvec$ linearly predicts cell count, a
different pair of variables from the $\rho(n_{\text{cells}}, y) = +0.705$ association of
Sec.~\ref{sec:endpoint-depth}; the two are reported separately and are not composed into a
mechanism in either direction. And this is a train\slash validation-only supporting
diagnostic: it neither resolves nor bears on H-FM1--H-FM4, whose statements stand as reported
in App.~\ref{app:hypotheses}.

\paragraph{Target-gene structure in the cell-count target.}
This sharpens the clustering limitation noted in Sec.~\ref{sec:discussion} and
App.~\ref{app:uncertainty}. Treating $\log(1 + n_{\text{cells}})$ as the response and
applying a one-way method-of-moments variance decomposition to VCC training and validation
rows separately: grouping by target gene gives an intraclass correlation of $0.86$ on
training ($140$ groups) and $0.82$ on validation ($49$ groups), while grouping by the
recorded batch gives small negative estimates on both ($-0.001$ and $-0.011$, over $48$
groups each). The batch estimates are computed from $N_{\text{batches}} = 48$ groups; when
the number of groups is limited, the between-batch mean square and the resulting
method-of-moments between-group component may be unstable. Cell count is therefore strongly clustered by target gene, with no positive
one-way clustering along batch.

Four qualifications are mandatory. The two intraclass correlations are separate one-way
estimates, mutually confounded to an unquantified degree; they are not components of a
joint decomposition and do not sum to a total. The associated design-effect figures
(roughly $34.4$--$39.3$ for target gene) are conventional approximations, not literal effective
sample sizes, and a negative estimate mechanically drives the corresponding figure below
one. Nothing here concerns the VCC test split. And nothing here shows that the outcome $y$
is itself target-gene clustered, that target gene is the true statistical unit, or that the
bootstrap coverage of App.~\ref{app:uncertainty} is thereby validated: the pre-specified
row-level interval was retained rather than replaced, as stated in
Sec.~\ref{sec:discussion}.

\section{Uncertainty quantification}
\label{app:uncertainty}

Two distinct resampling procedures are used in this paper. They are not
interchangeable, and we set them out separately.

\paragraph{In-distribution comparison on VCC (Sec.~\ref{sec:informativeness}).}
The statistic is the pooled $\Delta\rsq$ between two arms on the frozen VCC test
split. The resampling unit is the test row. The procedure is an independent row
bootstrap with $B = 10{,}000$ resamples and seed $42$; one resample index is
applied to \emph{both} arms in each replicate, so the comparison is paired, and
the three head-training seeds are averaged \emph{inside} each replicate.
Intervals are percentile intervals at the $2.5$th and $97.5$th points. No
clustering is used. For $\Delta\rsq(\zvec - C_1)$ this gives a point estimate of
$+0.1645$, a bootstrap median of $+0.1649$, a $95\%$ interval of
$[+0.1375, +0.1920]$, and sign-stability across every resample.

\paragraph{External transfer (Sec.~\ref{sec:transfer}).}
The statistic is the Spearman correlation between prediction and outcome, and
paired differences in that correlation between arms. The resampling unit is the
target-gene cluster. Each screen contains $1{,}811$ target-gene clusters, but these are
distinct screen-specific gene sets of equal cardinality, not a shared set. Constructs
sharing a target gene are resampled together. The procedure is a paired cluster
bootstrap with $B = 10{,}000$ resamples and seeds $20260901$ for RPE1 and
$20260902$ for K562-essential, with percentile intervals at the $2.5$th and
$97.5$th points; the same cluster resample is applied to both arms of a contrast
within each replicate, and head seeds are averaged \emph{outside} the bootstrap.

\paragraph{These are not the same procedure.}
The two differ in statistic---pooled $\Delta\rsq$ against Spearman
$\Delta\rho$---in resampling unit, row against target-gene cluster; in whether
clustering is applied at all; in seed; and in whether head seeds are averaged
inside or outside the resample. They share only the number of resamples and the
percentile method. Intervals from the two settings should not be read as though
produced by a common procedure.

\paragraph{Secondary magnitude correlations (Sec.~\ref{sec:max-response}).}
The count-adjusted partial Spearman correlations of $\max_j|x_j|$ with $y$ are
reported without intervals. No interval procedure was pre-specified for them and
none is persisted, so we report the point values and do not attach uncertainty
after the fact. The pre-specified robustness check for these correlations was a
within-quintile stratification of cell count; it was executed, and the
association is positive within every quintile on all three screens.

\section{Held-out data access ledger}
\label{app:test-ledger}

\paragraph{The VCC test split.}
Exactly one scientific evaluation of the frozen VCC test split was performed: the
frozen test table, produced after head selection had been written to disk. Two
further accesses are on the record and we list them rather than fold them into a
single count---a reproduction-gate run before the freeze, which did not complete,
and one post-hoc diagnostic access described in the session record. Neither
contributed a reported number, and every phase after the specification lock
records the test split as untouched. A single unqualified access count is not
derivable from the durable record, and we do not assert one.

\paragraph{The external screens.}
Under the locked protocol, external outcomes were to remain held out until final
evaluation. The surviving local record positively locates the first documented
external-outcome access at that point, while not excluding undocumented earlier access. Head selection
was frozen on one day; external predictions were materialized blind, and the
primary result persisted, on the next; and the first documented external-outcome access
follows both. Every prior session record states that the external outcome had not
been read.

\paragraph{Amendment chronology.}
Sixteen numbered amendments to the specification exist---fifteen as files, since A5 is a
fileless editorial event---and we separate them by timing rather than
describing the record with a single claim. A1 through A4 were authored and recorded between
2026-08-11 and 2026-08-13, each before the measurements it governs existed and before the
first positively located external-outcome access; A1's substance is supported from
2026-08-11, while its current text reflects a recorded edit of 2026-08-13 that added a
classification note without altering substance. They add pre-registered secondaries and
sensitivities, and none adds, removes, or reseeds a control; they first entered version
control later, on 2026-08-25. A5 and A6 were recorded after external unblinding and are
editorial: A5 rewords a paragraph of the introduction and A6 replaces four occurrences of one
phrase, and neither changes a hypothesis, a control, a seed, an endpoint, a statistic, or a
reported value. A5 is a fileless governance event with no artifact; A6 first entered version
control on 2026-08-26. A7 was also recorded after unblinding, entering version control on
2026-08-26, and is not editorial: it records the execution of the pre-registered non-overlap
sensitivity described above, states that the analysis was specified before unblinding but run
after the primary was persisted, and reports its outcome in full. A8 was recorded on
2026-08-27 and remained outside version control until the documentation pass accompanying this revision;
it is record-keeping, closing two documentation items the audit record had left open---the
eligibility reconciliation behind the split counts of Sec.~\ref{sec:setup}, together with the
universe against which the non-overlap exclusion was defined, and an encoder-width
inconsistency carried by an earlier draft and since corrected. A9 is a retrospective
governance and reporting reconciliation recorded with this revision. Neither changes a
hypothesis, control, seed, endpoint, statistic, reported value, or conclusion. A10, A11 and
A12 are later still, all recorded on 2026-08-29, well after external unblinding: they fix
the frozen preprocessing state for post-hoc work and govern the two supporting diagnostics
of App.~\ref{app:post-unblinding}. Each was locked before its own measurement existed, but
none belongs to the original pre-result specification, and none changes a hypothesis,
control, seed, endpoint, or reported value. A13, recorded after external evaluation, reconciled
the A3 RPE1 population description and restored previously omitted pre-specified secondary
reporting without changing any frozen primary result. A14, also post-unblinding, records the
final reporting-completion corrections identified by release audit, including full A1 secondary
reporting and reconciliation of the A4 secondary descriptions. A15, also post-unblinding, is a
wording reconciliation: it aligns chronology phrasing to the documented-access evidence bound and
changes no value.

\paragraph{A gap in the amendment token taxonomy.}
The change-control clause enumerates three classification tokens but never defines them,
and none of the three fits an additive pre-registered sensitivity. A2 and A3 therefore
carry the token \textsc{sensitivity\_addition}, which is not among the three, and A4
records the mismatch in place of a token. Each of the three was recorded before the
measurements it governs. We report this as a formal gap in the taxonomy; it does not bear
on what was run, when it was run, or any reported value.

\paragraph{No external fitting.}
Training and head selection used VCC labels alone. All preprocessing statistics
were fit on the VCC training split and applied frozen. No external calibration,
label-derived scaling, screen-specific refit, per-screen head choice, or
target-side tuning of any kind was performed.

\section{Chronology evidence}
\label{app:chronology-evidence}

The ordering of the specification, its amendments, model and head selection, and the
external unblinding is reconstructed from records local to the authoring machine. In the
surviving local evidence, amendments A1 through A4 precede the first positively located
external-outcome access: the latest of the four is supported from 2026-08-13T05:37:46Z,
and the first positively located external-outcome access lies between
2026-08-13T06:15:09.591Z and 2026-08-13T06:15:35.490Z, a margin of approximately
37~minutes 24~seconds. A1's substance is supported from 2026-08-11T19:09:21.979Z, while
its current text reflects a recorded edit of 2026-08-13T00:59:34.920Z that added a
classification note without altering substance.

The evidence classes are contemporaneous machine-written session records, filesystem
creation metadata, and process-written run artifacts. These are distinct local evidence
mechanisms that corroborate one another, but they are local to the authoring machine: they
are not externally timestamped, not cryptographically signed, and not maintained in an
append-only ledger, and they do not exclude an undocumented earlier access. Pinned SHA-256
digests cover the frozen foundation-model artifacts and the two harmonized external endpoint
files; the specification, lock and amendment chain do not carry an equivalent externally
anchored digest history.

Two gaps are disclosed rather than resolved. The session records provide substantial but
incomplete coverage of the experimental window: a 28.04-hour interval spanning 2026-08-12 has
no session-record entries, and the head-selection freeze falls inside it, resting instead on
filesystem metadata and the process-written phase-1 execution log, whose internal write order
places the freeze record before every test entry. Separately, no positive
\texttt{EXTERNAL\_Y\_READ = YES} runtime marker exists anywhere in the record, so the
unblinding moment is located by the surrounding artifacts rather than by the project's own
declared marker.

The counterfactual exclusion counts of App.~\ref{app:hypotheses} were reconstructed under a
frozen metadata-only procedure from the pre-specified metadata and set definitions, after
first reproducing the historical $43$ and $33$ construct overlap counts. The matching rule was
recovered from the executed analysis code rather than chosen, the procedure was fixed before
execution and run once, and no model was run and no external outcome column was accessed. The
frozen script has SHA-256 \texttt{9ed9deb1bd9123af4b136fd658c737f110f50102b9cca165e1e1ead548ee1ca9}.
This is a provenance reconstruction, not a proof and not a scientific validation.

\section{Locked evaluation protocol timeline}
\label{app:protocol-figure}
\begin{figure}[t]
  \centering
  \includegraphics[width=\linewidth]{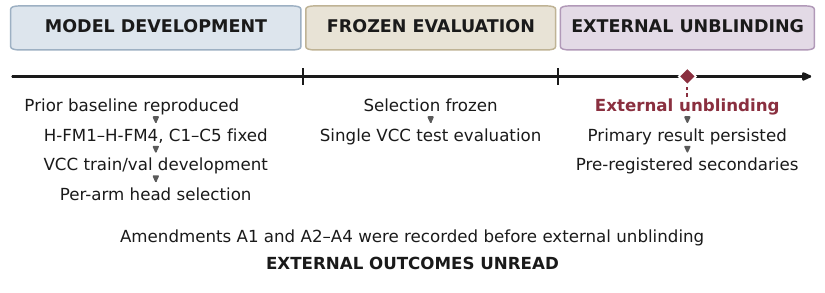}
  \caption{Locked, pre-registered evaluation protocol. Model and head selection are
  frozen on VCC before a single test evaluation; under the locked protocol the external
  outcome labels were to be held out until final evaluation; amendments that govern an
  analysis precede its results, and later amendments are recorded as dated documents. The
  in-distribution gate (H-FM1) precedes and conditions the zero-shot transfer reading
  (H-FM2). The figure's label \textsf{EXTERNAL OUTCOMES UNREAD} states that protocol
  requirement. The surviving local record positively locates the first documented
  external-outcome access at final evaluation and does not exclude undocumented earlier
  access (App.~\ref{app:chronology-evidence}). The \textsf{pre-registered secondaries} stage
  shown here is reported in App.~\ref{app:secondaries}.}
  \label{fig:protocol}
\end{figure}

Figure~\ref{fig:protocol} is the visual companion to the locked protocol of
Sec.~\ref{sec:protocol}: model and head selection are frozen on VCC before a single
test evaluation, the locked protocol required the external outcome labels to be held out
until final evaluation,
amendments that govern an analysis precede its results, and later amendments are
recorded as dated documents.

\section{Pre-specified hypotheses for the strong-baseline audit}
\label{app:phase11a-prereg}

\begin{table}[t]
\centering
\caption[Strong classical baselines on VCC]{Strong classical baselines on the VCC held-out target-gene test set (mean $\pm$ std, three seeds; Ridge and ElasticNet are deterministic at the displayed precision). Removing row-level magnitude (\textsc{dir}, $\xvec/\|\xvec\|_2$) sharply reduces performance, while concatenating $\xvec$ with the four magnitude statistics (\textsc{x+m}) enables tree ensembles to substantially outperform \magspsm{}. Bold marks the best value within each feature column.}
\label{tab:strong_baselines}
\footnotesize
\setlength{\tabcolsep}{4pt}
\begin{tabular}{lcccc}
\toprule
Method & $\xvec$ only & Direction only & Magnitude only & $\xvec$ + magnitude \\
\midrule
Ridge                & $-0.306$ & $-0.319$ & $\mathbf{+0.187}$ & $-0.107$ \\
ElasticNet           & $-0.247$ & $-0.252$ & $+0.175$ & $-0.037$ \\
Random Forest        & $+0.090{\scriptstyle\,\pm 0.004}$ & $-0.062{\scriptstyle\,\pm 0.003}$ & $-0.042{\scriptstyle\,\pm 0.005}$ & $\mathbf{+0.370{\scriptstyle\,\pm 0.004}}$ \\
HistGradientBoosting & $\mathbf{+0.190{\scriptstyle\,\pm 0.006}}$ & $\mathbf{+0.062{\scriptstyle\,\pm 0.008}}$ & $+0.054{\scriptstyle\,\pm 0.047}$ & $+0.365{\scriptstyle\,\pm 0.008}$ \\
$k$-NN ($k{=}10$)    & $-1.124$ & $-2.256$ & $-0.041$ & $-1.000$ \\
\midrule
\spsm{} (deep)        & \multicolumn{4}{c}{$+0.082 \pm 0.010$ \quad(no magnitude features)} \\
\magspsm{} (deep, ours) & \multicolumn{4}{c}{$+0.225 \pm 0.004$ \quad(deep proof-of-concept; uses $[\xvec;\,\magmaptilde]$)} \\
\maglinear{}          & \multicolumn{4}{c}{$+0.208$ \quad(four-feature linear regression on $\magmaptilde$)} \\
\bottomrule
\end{tabular}
\end{table}

\label{sec:strong-baselines}
We test whether the magnitude effect is specific to the deep \spsm{}
architecture or a more general property of held-out target-gene
prediction. We evaluated five classical methods (Ridge, ElasticNet,
Random Forest, HistGradientBoosting, $k$-NN) on four feature sets: the
full $\xvec$, the row-normalized direction $\xvec/\|\xvec\|_2$
(\textsc{direction-only}), $\magmaptilde$ alone (\textsc{mag-only}),
and the concatenation $[\xvec;\,\magmaptilde]$ (\textsc{x+mag}). All
preprocessing used training-split statistics; the four magnitude
features are pre-specified before any model training. Five hypotheses
were pre-specified before the experiment was run; we report all five
verdicts below.

The result (Table~\ref{tab:strong_baselines}) reframes the
interpretation of the deep model.
\emph{(i) Four magnitude scalars match the full input.} Ridge on
\textsc{mag-only} reaches $+0.187$, only $0.003$ below the strongest
classical model on $\xvec$ (HistGradientBoosting,
$+0.190 \pm 0.006$); \maglinear{} reaches $+0.208$. Four scalars carry
approximately as much held-out predictive signal as the full
$2{,}000$-dimensional vector under this split.
\emph{(ii) Removing magnitude collapses performance.} Row-normalizing
$\xvec$ reduces the strongest classical model from $+0.190$ to
$+0.062 \pm 0.008$. All linear methods become strongly negative on
\textsc{direction-only}.
\emph{(iii) Combining the two outperforms the deep encoder.} Random
Forest on $[\xvec;\,\magmaptilde]$ reaches $+0.370 \pm 0.004$ and
HistGradientBoosting reaches $+0.365 \pm 0.008$, both substantially
exceeding our deep \magspsm{} encoder ($+0.225 \pm 0.004$). Direction
combined with explicit magnitude adds $\sim 0.18\,\rsq$ that tree
ensembles, but not our deep encoder, exploit. That a low-dimensional
response-magnitude summary is recoverable by classical models here is therefore
not specific to a single MLP-encoder configuration; whether attention-based,
set-equivariant, or foundation-model encoders behave similarly is the question
the locked foundation-model evaluation takes up.
The Random Forest result on $[\xvec;\,\magmaptilde]$ is robust across
reasonable RF hyperparameter choices (App.~\ref{app:rf-sweep}).

The five hypotheses below were written in the experiment report before
the corresponding scripts were run; we have no formal external
preregistration, but each hypothesis is dated in the project log
preceding the run. We report each verdict honestly.

\begin{enumerate}[itemsep=2pt,topsep=2pt,leftmargin=*]
\item[H1] Classical models on $\xvec$ alone will not clearly beat
\maglinear{}. \emph{Verdict: supported.} Best $\xvec$-only model
(HistGradientBoosting, $\rsq = +0.190$) is below \maglinear{}
($+0.208$) and within $0.003$ of Ridge on \textsc{mag-only}
($+0.187$); no $\xvec$-only classical model exceeds \maglinear{} by a
margin larger than seed noise.
\item[H2] \textsc{direction-only} will perform worse than $\xvec$-only
and \textsc{mag-only}. \emph{Verdict: supported.} Best
\textsc{direction-only} ($+0.062$) is well below both anchors.
\item[H3] Linear methods on \textsc{mag-only} will approximately
reproduce \maglinear{}. \emph{Verdict: supported.} Ridge on
\textsc{mag-only} reaches $+0.187$ ($\geq +0.180$ pre-specified
threshold).
\item[H4] Tree ensembles on \textsc{x+mag} may match or beat
\magspsm{}. \emph{Verdict: clearly beats.} A row-level paired bootstrap
gives Random Forest on $[\xvec;\,\magmaptilde]$ over \magspsm{} a margin
of $\Delta\rsq = +0.145$ ($95\%$ CI $[+0.127, +0.163]$, sign-stable in
$100\%$ of resamples).
\item[H5] If a non-deep model on \textsc{x+mag} clearly beats
\magspsm{}, the paper should be reframed as diagnostic discovery rather
than method paper. \emph{Verdict: triggered.} The reframing applied to
this paper is the consequence of H5.
\end{enumerate}

\section{Per-row alignment: full population-level argument}
\label{app:per-row-alignment}

In the population view, let $M^\star$ be an auxiliary random vector
with the same marginal distribution as $\magmaptilde(X)$ but
independent of $(X, Y)$. Then the Bayes predictor with access to
$(X, M^\star)$ coincides with the Bayes predictor with access to $X$
alone:
\[
\Eb[Y \mid X, M^\star] \;=\; \Eb[Y \mid X].
\]
By the optimality of conditional expectation under squared loss, no measurable predictor using
$(X, M^\star)$ can improve over the Bayes risk achieved by
$\Eb[Y \mid X]$ in squared loss. This yields two consequences we test
directly: (a)~replacing $\magmaptilde(\xvec_i)$ by four i.i.d.\
$\mathcal{N}(0, I_4)$ features makes the augmenting block independent
of $(\xvec_i, Y_i)$ by construction; (b)~under a row-shuffled
augmentation
$\xvec_i^{\pi} = [\,\xvec_i;\, \magmaptilde(\xvec_{\pi(i)})\,]$ with
$\pi$ a uniformly random permutation independent of the data, the
augmenting block preserves the marginal distribution of
$\magmaptilde$ but no longer carries row-specific information about
$Y_i$. The shuffled-magnitude and random-feature controls of
App.~\ref{sec:causal} are finite-sample diagnostics of this principle:
they preserve dimensionality and marginal scale while breaking
row-specific alignment with the label. A systematic $\rsq$ gain over both
controls therefore supports attributing the contribution of the four scalars to
their per-row dependence on $\xvec_i$ rather than to added input dimensionality,
scale, or noise diversity. We do not promote this observation to a
formal theorem; its role is to motivate the two pre-specified
experimental controls (row-shuffling and i.i.d.\ Gaussian features)
reported in App.~\ref{sec:causal}.

\paragraph{Per-row alignment: empirical controls (Table~\ref{tab:causal}).}
We instantiated three additional arms at seed~$0$ (otherwise identical
training): (i) \textsc{baseline-mse}: \spsm{} V4 with no magnitude
features ($\xvec^{+} = \xvec$); (ii) \textsc{shuffled-magnitude}:
\magspsm{} but with per-row correspondence between $\xvec_i$ and
$\magmaptilde$ broken by a uniformly random permutation $\pi$; (iii)
\textsc{random-feature}: \magspsm{} but with the magnitude block
replaced by four i.i.d.\ $\mathcal{N}(0,1)$ features. The result
(Table~\ref{tab:causal}) is unambiguous: both controls return
performance to baseline within seed noise. Shuffling removes the gain
entirely ($\Delta = -0.151$ vs.\ \magspsm{}); four random Gaussian
features fail to recover any gain. Under shuffling,
$\magmaptilde(\xvec_{\pi(i)}) \indep (\xvec_i, y_i)$ in the sample
limit, so the augmenting block carries no information about $y_i$.

\begin{table}[t]
\centering
\caption{Per-row alignment controls (seed 0). Both shuffled-magnitude (per-row
alignment broken) and four random Gaussian features (independent of
input by construction) return performance to baseline within seed
noise, consistent with the population-level argument of this
appendix.}
\label{tab:causal}
\small
\setlength{\tabcolsep}{4pt}
\begin{tabular}{lcccc}
\toprule
Arm & Test $\rsq$ & Spearman & Q1 bias & Q5 bias \\
\midrule
\textsc{baseline-mse} (no $\magmaptilde$)             & $+0.0768$ & $+0.399$ & $+1.061$ & $-0.735$ \\
\magspsm{} (4 magnitude features)                     & $\mathbf{+0.2224}$ & $\mathbf{+0.576}$ & $+0.918$ & $-0.580$ \\
\textsc{shuffled-magnitude} (per-row $\pi$ random)    & $+0.0712$ & $+0.402$ & $+1.191$ & $-0.592$ \\
\textsc{random-feature} (four $\mathcal{N}(0,1)$)     & $+0.0869$ & $+0.377$ & $+1.144$ & $-0.660$ \\
\bottomrule
\end{tabular}
\end{table}

\section{MAG-Anchor: full architecture, hyperparameters, and controls}
\label{app:mag-anchor-full}
\label{sec:mag_anchor_results}

MAG-Anchor is a secondary proof-of-concept deep method, not a state-of-the-art
in-distribution predictor, and was not reproduced under the locked revision harness; its
previously reported performance figures are not carried into this revision. Its architecture and
the pre-specified per-row alignment controls (eval-time row shuffling, train-time row
shuffling, and random-Gaussian replacement of $\magmaptilde$, each of which collapses the
gain, so the improvement depends on correctly aligned per-row magnitude features rather
than on added capacity) are given below.
Table~\ref{tab:main_results_with_mag_anchor} lists the in-distribution results that were
reproduced under the locked harness, with MAG-Anchor named but its previously reported figures
withheld.

\begin{table}[t]
\centering
\caption[VCC in-distribution results]{VCC in-distribution held-out target-gene results, reproduced under the locked revision harness (mean $\pm$ std across seeds, $\mathrm{ddof}{=}1$, where the model is stochastic). MAG-Anchor was not reproduced under the locked revision harness and is described qualitatively (App.~\ref{app:mag-anchor-full}); its previously reported performance figures are not carried into this revision.}
\label{tab:main_results_with_mag_anchor}
\footnotesize
\setlength{\tabcolsep}{6pt}
\begin{tabular}{@{}lc@{}}
\toprule
Model & VCC $R^2$ \\
\midrule
SPSM baseline & $+0.082 \pm 0.010$ \\
MAG-SPSM & $+0.225 \pm 0.004$ \\
MAG-LINEAR & $+0.208$ \\
RF (x + mag) & $+0.370 \pm 0.004$ \\
\bottomrule
\end{tabular}
\end{table}

\paragraph{Architecture.}
MAG-Anchor decomposes the predicted perturbation effect into a direct
magnitude anchor and a gated residual correction:
\begin{equation}
\hat{y}
=
g_\phi\!\left(\magmaptilde(\xvec)\right)
+
\alpha\, h_\theta\!\left([\xvec;\magmaptilde(\xvec)]\right),
\label{eq:mag_anchor}
\end{equation}
where $\magmaptilde(\xvec)\in\Rb^4$ is the standardized magnitude
vector defined in Eq.~\ref{eq:mag-features}. The anchor branch
$g_\phi(\magmaptilde) = \wvec^{\top}\magmaptilde + b$ is a single
linear layer with $\wvec \in \Rb^{4}$, $b \in \Rb$;
its weights are warm-started from a sklearn \texttt{LinearRegression}
fitted on $(\magmaptilde, y)$ and remain trainable thereafter (no
freezing). The residual branch $h_\theta$ is a small MLP with layers
$d_{\mathrm{in}} \to 256 \to 128 \to 1$, ReLU activations, and
architectural dropout $p{=}0.1$ after each ReLU; no LayerNorm or
BatchNorm is used. This dropout setting applies only to the
MAG-Anchor residual branch; the \spsm{}/\magspsm{} backbone
(App.~\ref{app:spsm}) uses dropout $p{=}0$. For the headline
MAGAnchor-xmag the residual input
is $[\xvec;\,\magmaptilde] \in \Rb^{2004}$ ($d_{\mathrm{in}}=2004$);
for the MAGAnchor-x ablation the residual input is $\xvec$ alone
($d_{\mathrm{in}}=2000$). The scalar gate $\alpha$ is parameterized
as $\alpha = \mathrm{softplus}(\rho)$ with
$\rho_{\mathrm{init}} = \ln(\mathrm{e}^{0.1}-1)$, so
$\alpha_{\mathrm{init}} = 0.1$ exactly and $\alpha \in (0, \infty)$
throughout training. This design preserves a direct pathway from
magnitude features to the output while allowing the residual branch
to learn corrections from both the expression vector and the
magnitude summary.

\paragraph{Training hyperparameters.}
Unless otherwise stated, MAG-Anchor uses a trainable anchor, no
residual penalty, Adam with learning rate $10^{-3}$ and weight
decay $10^{-5}$, batch size $64$, $25$ training epochs, and
checkpoint selection by validation $\rsq$. Frozen-anchor and
residual-penalty variants were evaluated only as sensitivity checks; they were not
reproduced under the locked revision harness and no value from them is reported.

\paragraph{Per-row alignment is load-bearing (controls).}
The earlier MAG-Anchor analysis pre-specified three alignment controls:
permuting $\magmaptilde(\xvec)$ at evaluation, permuting it during training,
and replacing it with Gaussian features. They were recorded as collapsing the
gain. Because MAG-Anchor was not reproduced under the locked revision harness,
we retain this only as historical context and do not use legacy Replogle behavior
as current evidence.

\section{Classical-baseline hyperparameters}
\label{app:classical}

\begin{itemize}[itemsep=2pt,topsep=2pt,leftmargin=*]
\item Ridge: \texttt{Ridge(alpha=1.0)}.
\item ElasticNet: \texttt{ElasticNet(alpha=0.001, l1\_ratio=0.5,
max\_iter=10000)}.
\item Random Forest: \texttt{RandomForestRegressor(n\_estimators=200,
max\_depth=20, min\_samples\_leaf=2, n\_jobs=8, random\_state=seed)}.
\item HistGradientBoosting (XGBoost fallback): default sklearn
\texttt{HistGradient\allowbreak Boosting\allowbreak Regressor} with
\texttt{max\_iter=500}, \texttt{learning\_rate=0.05},
\texttt{max\_depth=6}, \texttt{random\_state=seed}, early stopping
on validation with patience 20.
\item $k$-NN: \texttt{KNeighborsRegressor(n\_neighbors=10,
weights="distance")}.
\end{itemize}
For linear models and $k$-NN, features are standardized using
training-split mean/std exclusively; for tree models, raw features are
used (with the four magnitude features $z$-scored regardless of model).

\section{\spsm{} V4 backbone and loss formulas}
\label{app:spsm}

The \spsm{} V4 backbone is a three-layer MLP encoder with hidden
width $256$, ReLU activations, and LayerNorm; architectural dropout
is $p=0$ (no active dropout regularization). The encoder produces a
$256$-dimensional representation $\zvec_i = f_\theta(\xvec_i^{+})$.
The total objective combines a prediction loss with two SPSM
regularizers,
$\mathcal{L}_{\mathrm{total}} = \mathcal{L}_{\mathrm{MSE}} +
\lambda_{\mathrm{stab}}\,\mathcal{L}_{\mathrm{stab}} +
\lambda_{\mathrm{rel}}\,\mathcal{L}_{\mathrm{rel}}$, with
$\lambda_{\mathrm{stab}} = \lambda_{\mathrm{rel}} = 0.5$, both
inherited from the SPSM V4 backbone unchanged.

\paragraph{Stability loss.}
The stability term is a Gaussian-kernel maximum-mean-discrepancy
(MMD) penalty on encoder features across the experimental Flex
batches that co-occur within the same SGD minibatch; no stochastic
input perturbation, dropout-perturbed forward pass, or input noising
is used. Concretely, for an SGD minibatch $\mathcal{I}$ of size
$B = 64$, let $\mathcal{B}(\mathcal{I}) = \{b_i : i \in \mathcal{I}\}$
denote the set of experimental Flex batches represented in
$\mathcal{I}$, and let
$Z_b(\mathcal{I}) = \{\zvec_i : i \in \mathcal{I},\, b_i = b\}$
collect the encoder outputs from rows of batch $b$. The stability
loss averages the squared MMD across all unordered pairs of
co-represented batches:
\[
\mathcal{L}_{\mathrm{stab}}
\;=\;
\frac{2}{|\mathcal{B}(\mathcal{I})|\,(|\mathcal{B}(\mathcal{I})|-1)}
\!\!\sum_{\substack{b,b' \in \mathcal{B}(\mathcal{I}) \\ b<b'}}
\mathrm{MMD}^2_{\kappa}\!\left(Z_b(\mathcal{I}),\, Z_{b'}(\mathcal{I})\right),
\]
where $\kappa$ is a Gaussian kernel whose bandwidth is set by the
median heuristic on the in-minibatch pairwise distances, computed
with gradients disabled. For each SGD minibatch $\mathcal{I}$, $\mathcal{B}(\mathcal{I})$
denotes the experimental batches represented in that minibatch. The
sampler constructs context-stratified minibatches so that each
represented batch contributes enough rows for the MMD pairs in the stability loss below to be well-defined.

\paragraph{Relation loss.}
The relation loss is a soft-Spearman ranking penalty applied to the
scalar predictions $\hat{y}$ across pairs of experimental batches
that co-occur in the same minibatch; it does not use a learnable
projection on $\zvec$ and does not enter the labels $y_i$ (labels
enter only through $\mathcal{L}_{\mathrm{MSE}}$). For an SGD
minibatch $\mathcal{I}$, we construct a small set $\mathcal{P}$ of
matched context-pairs by iterating over unordered pairs of distinct
Flex batches $(b, b') \in \mathcal{B}(\mathcal{I}) \times
\mathcal{B}(\mathcal{I})$ and, for each pair, taking the first
applicable of three tiers: tier~1 (weight $w = 1.00$) matches rows
with the same target-gene identifier shared between $b$ and $b'$;
tier~2 (weight $w = 0.50$) matches rows with the same auxiliary
cluster index, paired greedily; tier~3 (weight $w = 0.25$) takes
$k{=}1$ nearest neighbors in an auxiliary $u$-space. On VCC, only
tier~1 contributes informative matches, so in practice $\mathcal{P}$
contains pairs of Flex batches with at least one shared target gene,
aligned by target gene. Each matched pair $p \in \mathcal{P}$ yields
two aligned prediction vectors
$\hat{\boldsymbol{y}}^{(p)}_a, \hat{\boldsymbol{y}}^{(p)}_b
\in \Rb^{m_p}$ and a weight $w_p$. With soft ranks
$\widetilde{r}_k(\boldsymbol{v}; \tau) = \sum_{\ell \neq k}
\sigmoid\!\bigl((v_k - v_\ell)/\tau\bigr)$, the soft-Spearman
correlation
\[
\widetilde{\rho}(\boldsymbol{a}, \boldsymbol{b}; \tau) \;=\;
\frac{\sum_k (\widetilde{r}_k(\boldsymbol{a}) - \overline{\widetilde{r}}(\boldsymbol{a}))
              (\widetilde{r}_k(\boldsymbol{b}) - \overline{\widetilde{r}}(\boldsymbol{b}))}
     {\sqrt{\sum_k (\widetilde{r}_k(\boldsymbol{a}) - \overline{\widetilde{r}}(\boldsymbol{a}))^2 \,
            \sum_k (\widetilde{r}_k(\boldsymbol{b}) - \overline{\widetilde{r}}(\boldsymbol{b}))^2}},
\]
the relation loss is
\[
\mathcal{L}_{\mathrm{rel}}
\;=\;
\frac{\sum_{p \in \mathcal{P}} w_p \,
      \bigl(1 - \widetilde{\rho}(\hat{\boldsymbol{y}}^{(p)}_a,
                                 \hat{\boldsymbol{y}}^{(p)}_b;\, \tau)\bigr)}
     {\sum_{p \in \mathcal{P}} w_p},
\qquad \tau = 0.1.
\]
$\mathcal{L}_{\mathrm{rel}}$ thus drives the model to produce a
consistent rank order of predictions across pairs of experimental
batches in the same minibatch.

\paragraph{Optimization.}
Adam at lr $10^{-3}$, weight decay $10^{-5}$, batch size $64$, $25$
epochs; HVG selection on the training split only; magnitude $z$-score
statistics from the training split only; model selection by
validation $\rsq$.

\begin{figure}[h]
  \centering
  \includegraphics[width=0.95\linewidth]{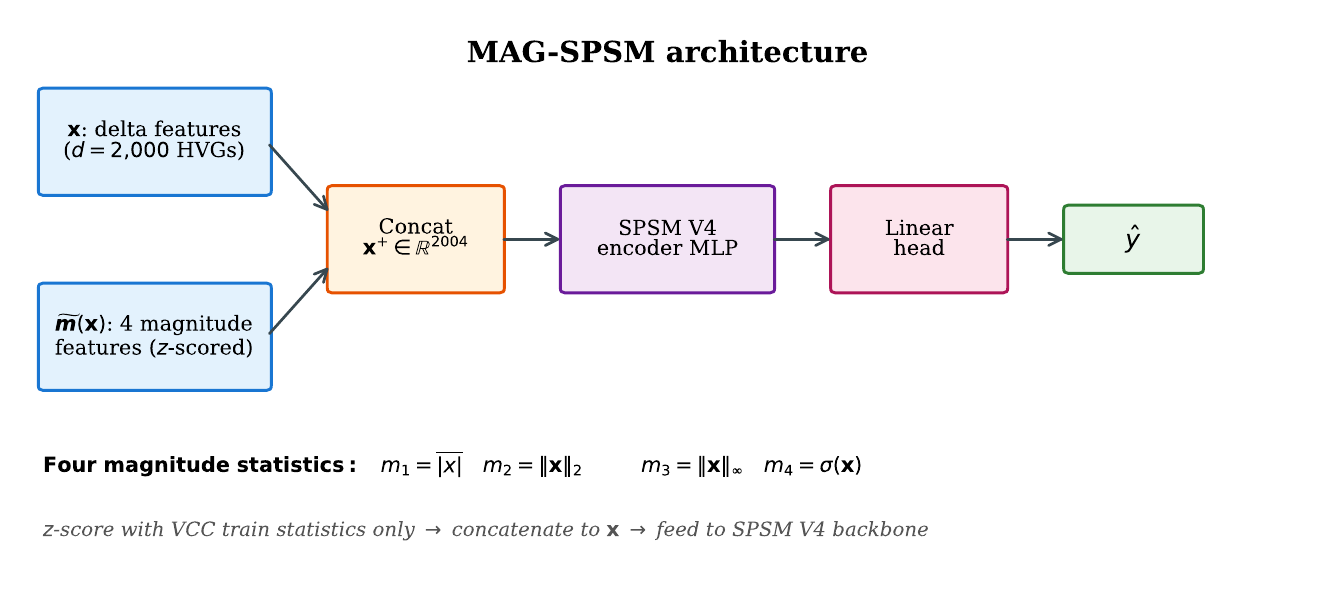}
  \caption[MAG-SPSM architecture]{\magspsm{} architecture. The expression delta vector $\xvec$
  ($d{=}2{,}000$ HVGs) is concatenated with the four $z$-scored
  magnitude features $\magmaptilde(\xvec) =
  (\overline{|x|},\, \|\xvec\|_2,\, \|\xvec\|_\infty,\,
  \sigma(\xvec))$ to form $\xvec^{+} \in \Rb^{2004}$, which is passed
  through the \spsm{} V4 encoder MLP (three hidden layers of width $256$,
  ReLU, LayerNorm, no architectural dropout) and a linear head. All
  $z$-score statistics use VCC training data exclusively.}
  \label{fig:method-overview}
\end{figure}

\paragraph{Architecture overview (visual).}
Figure~\ref{fig:method-overview} summarizes the \magspsm{} pipeline.
The four magnitude statistics $\magmap(\xvec)$ defined in
Eq.~\eqref{eq:mag-features} are $z$-scored using VCC training statistics
to obtain $\magmaptilde(\xvec) \in \Rb^{4}$, then concatenated with the
$d{=}2{,}000$-dimensional log1p delta input $\xvec$ to form
$\xvec^{+} \in \Rb^{2004}$, which is fed to the \spsm{} V4 backbone and
a linear head. The same backbone, the same losses, and the same
optimizer are used for the \spsm{} baseline (which receives only
$\xvec$) and for \magspsm{} (which receives $\xvec^{+}$); the only
difference is whether the four magnitude features are exposed to the
encoder.

\section{Magnitude statistics on Replogle}
\label{app:replogle-mag}

For the primary $2{,}000$-HVG transfer evaluation, each Replogle row
$\xvec^{(R)}$ is first embedded into the VCC HVG namespace by
zero-imputing the $570$ VCC-only coordinates missing from Replogle.
The magnitude statistics $\magmap(\xvec^{(R)})$ are then computed using
Eq.~\eqref{eq:mag-features} over this $2{,}000$-dimensional vector,
including the zero-imputed entries. In the HVG-coverage controls
(App.~\ref{app:hvg-control}), by contrast, models are trained and
evaluated directly in the $1{,}430$ shared-HVG subspace. We then $z$-score using \emph{VCC
training} statistics $\boldsymbol{\mu}_{\text{tr}},
\boldsymbol{\sigma}_{\text{tr}}$ (App.~\ref{sec:method}), preserving
the strict zero-shot constraint. The Replogle $z$-scored magnitudes
have empirical means roughly $+0.8\,\sigma_{\text{tr}}$ to
$+1.9\,\sigma_{\text{tr}}$ relative to the VCC training baseline, and
component-wise standard deviations $1.6\times$ to $2.4\times$ wider,
reflecting the larger feature scale of the Replogle release. These shifted
features lie outside VCC training support; we do not infer a model-specific OOD
mechanism or penalty from that observation.

\paragraph{Optimization, training schedule, and compute.}
All deep models (SPSM, MAG-SPSM, MAG-Anchor) use Adam at learning
rate $10^{-3}$ with weight decay $10^{-5}$, batch size $64$, and $25$
training epochs. The loss \eqref{eq:total-loss} applies to
\spsm{}/\magspsm{}; MAG-Anchor uses MSE on $\hat{y}$. HVG selection
and all $z$-score statistics use the VCC training split exclusively.
All deep experiments fit on a single TITAN~RTX 24\,GB GPU; classical
baselines were run on CPU. Classical baselines use
\texttt{scikit-learn} defaults (detailed in App.~\ref{app:classical}).

\section{Additional results: ablations, rank extension, and negative results}
\label{app:rank}

\paragraph{Single-feature and leave-one-out ablation of \magspsm{}.}
We performed a complete single-feature and leave-one-out (LOO) ablation
of the four magnitude features at seed~$0$
(Table~\ref{tab:ablation}). Across the four single-feature variants,
the VCC test $\rsq$ ranges from $+0.095$ (single $\|\xvec\|_{2}$) to
$+0.128$ (single $\overline{|\xvec|}$), all well below the full
4-feature \magspsm{} at $+0.222$. Across the four leave-one-out
variants, the test $\rsq$ ranges from $+0.137$ (drop $\max|\xvec|$) to
$+0.230$ (drop $\|\xvec\|_{2}$); the latter is the best LOO arm and is
within $0.008$ of the full 4-feature model, while removing
$\max|\xvec|$ costs $0.086$, indicating that $\max|\xvec|$ contributes
the most among the four magnitude features.

\begin{table}[t]
\centering
\caption[Magnitude feature ablation on VCC test]{Magnitude feature ablation on VCC test (seed 0). $\Delta$ vs full = R$^2$ minus full 4-feature MAG-SPSM. Bold marks the full four-feature model and the best leave-one-out variant.}
\label{tab:ablation}
\small
\begin{tabular}{lcccc}
\toprule
Variant & Test R$^2$ & $\Delta$ vs base & $\Delta$ vs full & Spearman \\
\midrule
SPSM baseline (no MAG) & $+0.0768$ & $+0.0000$ & $-0.1456$ & $+0.3986$ \\
\textbf{MAG-SPSM (full 4)} & $+0.2224$ & $+0.1456$ & $+0.0000$ & $+0.5763$ \\
only $m_1$ (mean $|x|$) & $+0.1278$ & $+0.0510$ & $-0.0946$ & $+0.4987$ \\
only $m_2$ ($\ell_2$) & $+0.0950$ & $+0.0181$ & $-0.1274$ & $+0.4557$ \\
only $m_3$ ($\max |x|$) & $+0.1269$ & $+0.0501$ & $-0.0955$ & $+0.4406$ \\
only $m_4$ (std) & $+0.1001$ & $+0.0233$ & $-0.1223$ & $+0.4531$ \\
drop $m_1$ & $+0.1986$ & $+0.1218$ & $-0.0238$ & $+0.5730$ \\
\textbf{drop $m_2$} & $+0.2303$ & $+0.1535$ & $+0.0079$ & $+0.5884$ \\
drop $m_3$ & $+0.1365$ & $+0.0597$ & $-0.0859$ & $+0.5084$ \\
drop $m_4$ & $+0.2084$ & $+0.1316$ & $-0.0140$ & $+0.5784$ \\
\bottomrule
\end{tabular}
\end{table}

\paragraph{\magrank{}: a small rank-margin extension of \magspsm{}.}
\magrank{} adds a margin-based pairwise rank loss
$\mathcal{L}_{\text{rank}} = \tfrac{1}{|\mathcal{P}|} \sum_{(i,j) \in
\mathcal{P}} \max(0,\, \alpha - \mathrm{sign}(y_i - y_j)\cdot
(\hat{y}_i - \hat{y}_j))$ over within-batch close-call pairs, with
$\alpha = 0.1$, in-batch close-call pairs dropped at
$\varepsilon_y = 0.05$, and weight $\lambda_{\text{rank}} = 0.1$. The
3-seed test $\rsq$ rises to $+0.2396 \pm 0.0078$, a paired improvement
over \magspsm{} of $+0.0142 \pm 0.0040$. Spearman is statistically
unchanged; we report \magrank{} as a small reproducible extension and
do not promote it as a contribution. The strong-baseline result of
App.~\ref{sec:strong-baselines} dominates this gain by an order of
magnitude.

\paragraph{Per-quintile bias.}
Relative to \spsm{}, \magspsm{} reduces the signed tail biases from $+1.093$
to $+0.918$ in Q1 and from $-0.696$ to $-0.580$ in Q5, consistent with the
diagnostic that magnitude features expand the prediction range, but substantial
residual calibration bias remains. The residual per-quintile bias
is not closed by \magspsm{}, motivating the strong-baseline audit
(App.~\ref{sec:strong-baselines}) and the cross-dataset experiment
(Sec.~\ref{sec:transfer}).

\begin{figure}[h]
  \centering
  \includegraphics[width=0.62\linewidth]{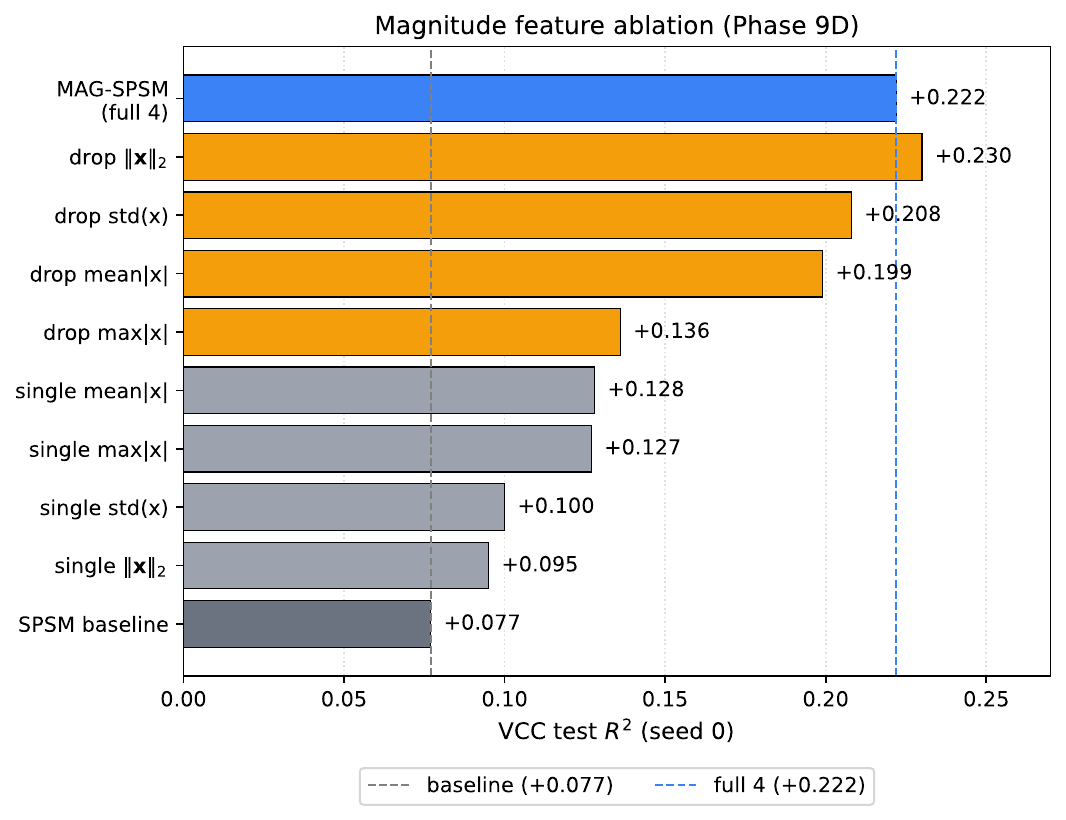}
  \caption{Single-feature and leave-one-out ablations of the four magnitude features.
  The four form a partially redundant low-dimensional summary: dropping
  $\|\xvec\|_2$ leaves performance essentially unchanged, whereas dropping
  $\max_j|x_j|$ costs the most.}
  \label{fig:ablation}
\end{figure}

The single-seed feature ablation is summarized in Fig.~\ref{fig:ablation}; it is
diagnostic rather than a primary multi-seed result.

Dataset-level split and scale summaries are provided in
Table~\ref{tab:dataset_statistics}.

\begin{table}[t]
\centering
\caption[Dataset statistics]{Dataset statistics for the raw releases and the earlier
distributed-endpoint analysis. Both screens use CRISPRi perturbation and differ in cell
line, biological-row unit, and feature normalization. \textbf{These are not the eligible
construct sets used by the locked harmonized transfer evaluation}, which retains
$1{,}932$ RPE1 and $1{,}903$ K562-essential constructs (Sec.~\ref{sec:transfer}), and the
gene-overlap row below counts a different quantity from the training overlap reported
there.}
\label{tab:dataset_statistics}
\footnotesize
\setlength{\tabcolsep}{4pt}
\begin{tabular}{lll}
\toprule
 & VCC & Replogle K562 (essential) \\
\midrule
Source & Virtual Cell Challenge & Replogle et al.\ 2022 (Cell) \\
Cell type & H1 hESC (stem) & K562 (CML) \\
Perturbation & CRISPRi (dCas9-KRAB) & CRISPRi (dCas9-KRAB) \\
Train / Val / Test rows & 5{,}740 / 2{,}052 / 3{,}972 & --- (held-out test only) \\
Non-NT rows used (test) & 3{,}972 & 2{,}176 \\
Unique target genes & 282 (all splits) & 2{,}057 \\
Train/Val/Test target-gene split & 140 / 49 / 93 (\textit{disjoint}) & --- \\
Feature dim ($d_x$ used) & 2{,}000 (HVG) & 8{,}563 full; 1{,}430 in HVG \\
Feature normalization & log1p $\Delta$ vs per-batch NT & pre-z-scored log1p change \\
Feature std (post-HVG) & $0.085$ & $0.161$ ($\sim$\,$1.9\times$) \\
Label & log1p(AD) per (batch, gene) & log1p(AD) per (pert, prom) \\
Label range & $[0,\,5.72]$ & $[0,\,8.71]$ \\
Label mean (std) & $4.23\,(0.81)$ & $3.99\,(2.75)$ \\
Per-row biological unit & (batch $\times$ gene) pseudobulk & (pert, prom) batch-aggregated \\
VCC target genes (all splits) also assayed & --- & 68 / 282 (24.1\%) \\
\bottomrule
\end{tabular}
\end{table}

\section{Endpoint provenance for the distributed Replogle column}
\label{app:endpoint-provenance}

\paragraph{Reconstruction attempt.}
We tested whether the Anderson--Darling column distributed with the
Replogle releases can be reproduced from the corresponding single-cell
data under the definition of Sec.~\ref{sec:setup}. Using RPE1, where
both the distributed column and single-cell counts are available, we
selected $30$ constructs by a deterministic rule---sorted by the stored
value, binned into ten equal-frequency strata, three per stratum---and
computed the two-sample Anderson--Darling statistic at each target
gene's own column under eight pre-declared variants: aggregation
(pooled across batches; per-batch then unweighted mean, count-weighted
mean, or median), transform (raw statistic; $\log1p(\max(0,\cdot))$),
input matrix (raw counts; normalized expression), and control sample
(non-targeting cells pooled globally; restricted to the same batch).

No variant reproduced the distributed values. The closest was per-batch
median on normalized expression with globally pooled controls, reaching
Pearson $0.335$ and Spearman $0.268$ with normalized RMSE $0.30$. We
classify the outcome as divergent: the distributed column is not a
target-gene Anderson--Darling effect size under any tested construction.

\paragraph{Evidence for a transcriptome-wide interpretation.}
The distributed values are integer-valued and bounded well below the
gene count in every screen ($[0, 6036]$, $[0, 5547]$, and $[0, 5528]$
for K562-essential, RPE1, and K562 genome-wide), which is consistent
with a per-perturbation count over genes. We therefore computed
transcriptome-wide per-gene Anderson--Darling statistics for ten
deterministically selected constructs and compared eight pre-declared
count definitions against the stored values: counts of genes with
statistic above zero, above $1.0$, or above $2.0$; counts with
$p < 0.05$ or $p < 0.01$, uncorrected or after Benjamini--Hochberg
correction; and a detection-count null.

All count candidates were rank-monotone with the stored column
($\rho = 0.96$--$0.98$). The Benjamini--Hochberg-corrected count at
$q < 0.05$ on normalized expression reached Pearson $0.986$; the
detection-count null was rejected, being constant at $8{,}749$. The
fitted slope was approximately $0.60$: our reconstruction runs roughly
$1.6\times$ the stored value. The exact upstream preprocessing,
significance threshold, control set, and tested gene universe therefore
remain unresolved, and we do not claim to have reconstructed the
column. We note that \texttt{scipy.stats.anderson\_ksamp} floors its
reported significance level at $0.001$ and caps it at $0.25$, which
makes any threshold-based count sensitive to implementation detail.

\paragraph{Association with the target-gene endpoint.}
Within each screen, the distributed column and the rebuilt target-gene
endpoint are only weakly associated at construct level: Spearman
$0.165$ $[0.118, 0.211]$ in RPE1, $0.301$ $[0.257, 0.343]$ in
K562-essential, and $0.342$ $[0.322, 0.362]$ at target level in the
K562 genome-wide screen. The two endpoints capture largely distinct
aspects of perturbation effect.

\section{Harmonized endpoint construction}
\label{app:harmonized-endpoint}

\paragraph{Protocol.}
The subsample sizes are the medians of the VCC distribution: $27$
perturbed cells per row and $788$ non-targeting cells per batch. For
each eligible construct we draw $27$ perturbed cells without
replacement from its pooled cells and $788$ non-targeting cells without
replacement from the pooled control set, compute the log1p
Anderson--Darling statistic at the target gene's column, and repeat $25$
times with pre-specified seeds. The endpoint is the median across
draws; we also record the mean, standard deviation, interquartile
range, and $5$th and $95$th percentiles. Quantiles use
\texttt{numpy.quantile} with linear interpolation throughout.

Eligibility requires at least $\max(20, 27)$ perturbed cells,
at least $\max(200, 788)$ pooled control cells, an unambiguous mapping
from target gene to exactly one feature column, and a verified
one-to-one join between the bulk and single-cell artifacts.

\paragraph{Coverage.}
Eligibility retains $1{,}932$ constructs for RPE1 and $1{,}903$ for
K562-essential. Exclusions are dominated by target genes absent from
the feature namespace rather than by insufficient cells. We report
retained counts rather than a retention fraction: the pre-eligibility
denominator is the output of an internal construct-selection step and
is not recoverable from a governed artifact, so a percentage would
carry a denominator we cannot source.

\paragraph{Resampling spread.}
The ratio of the median within-construct interquartile range to the
between-construct standard deviation is $0.174$ at construct level and
$0.184$ at target level for RPE1, and $0.236$ and $0.258$ for
K562-essential. Within-construct resampling variation is therefore
small relative to between-construct variation in both screens.

\paragraph{Pooled-scale comparison.}
A simple pooled construction using all eligible cells, without
subsampling, correlates with the matched-size endpoint at Spearman
$0.884$ $[0.871, 0.895]$ in RPE1 and $0.874$ $[0.860, 0.886]$ in
K562-essential. Rank agreement is strong but not exact: matching the
subsample size changes some rankings, which is why we fix it rather
than pooling.

\paragraph{Control-matching sensitivity.}
Restricting control cells to the same batch as the perturbed cells,
rather than pooling globally, is possible for $43$ constructs in RPE1
and $194$ in K562-essential under a minimum of ten perturbed and fifty
control cells per batch. Where computable, the batch-matched endpoint
correlates with the pooled-control endpoint at Spearman $0.910$ in RPE1
and $0.973$ in K562-essential. The two constructions therefore show high rank
agreement in the subsets where both are computable; we do not extrapolate that to an
equivalence across the full construct sets. We
note that the batch variable available in these releases is a technical
grouping whose status as an independent experimental replicate is not
established by the accompanying metadata.

\section{MAG-Anchor and external transfer}
\label{app:mag-anchor-transfer}

MAG-Anchor is evaluated in-distribution on VCC
(App.~\ref{sec:mag_anchor_results}) but does not appear in
Table~\ref{tab:harmonized_transfer}. The per-construct prediction
vectors required to evaluate it against a rebuilt endpoint were not
retained for the external screens; only aggregate metrics were. Since
our transfer evaluation re-scores stored predictions rather than
retraining, and no stored MAG-Anchor prediction exists to validate a
reconstruction against, we exclude it rather than introduce an
unvalidated refit into the comparison. This is an availability
constraint, not a selection based on outcome.

\section{Bootstrap intervals for the earlier classical comparisons}
\label{app:bootstrap-ci}

The intervals in this section belong to the classical-baseline analysis of the
earlier version of this work and are distinct from the resampling procedures that
govern the results of Sec.~\ref{sec:informativeness} and Sec.~\ref{sec:transfer},
which are specified in App.~\ref{app:uncertainty}. We
computed row-level paired bootstrap $95\%$ confidence intervals
($B = 10{,}000$ resamples, RNG seed $42$, percentile method) for three
pre-specified comparisons. The summary is in
Table~\ref{tab:bootstrap}. (C5)~Random Forest on
$[\xvec;\,\magmaptilde]$ versus \magspsm{} on VCC test pooled $\rsq$,
paired across the same three seeds: median $\Delta\rsq = +0.145$,
$95\%$ CI $[+0.127, +0.163]$, sign-stable in $100\%$ of resamples
(\textsc{strong}). (C7)~\maglinear{} versus HistGradientBoosting on
$\xvec$-only on VCC test $\rsq$ against the mean of three HistGB
seeds: median $\Delta\rsq = +0.012$, $95\%$ CI $[-0.009, +0.032]$
(\textsc{weak}). (C8)~\maglinear{} versus the \spsm{} baseline on
K562 Spearman $\rho$, row-resampled with no seed component: median
$\Delta\rho = +0.750$, $95\%$ CI $[+0.700, +0.800]$ (\textsc{strong}).

\begin{table}[h]
\centering
\small
\caption[Row-level paired bootstrap confidence intervals]{Row-level paired bootstrap $95\%$ confidence intervals for
three pre-specified comparisons. $B = 10{,}000$ resamples, RNG seed
$42$, percentile method. Sign-stability is the fraction of bootstrap
resamples in which the sign of the paired difference matches the
median.}
\label{tab:bootstrap}
\begin{tabular}{llccc}
\toprule
ID & Comparison & Metric & Median $\Delta$ & $95\%$ CI \\
\midrule
C5 & RF $[\xvec;\,\magmaptilde]$ vs.\ \magspsm{} (VCC) & $\rsq$ & $+0.145$ & $[+0.127, +0.163]$ \\
C7 & \maglinear{} vs.\ HistGB $\xvec$-only (VCC) & $\rsq$ & $+0.012$ & $[-0.009, +0.032]$ \\
C8 & \maglinear{} vs.\ \spsm{} baseline (K562) & $\rho$ & $+0.750$ & $[+0.700, +0.800]$ \\
\bottomrule
\end{tabular}
\end{table}

We do not report per-row bootstrap intervals for comparisons involving
MAG-Anchor or for the asymmetric three-seed-vs-one-seed
RF--\magspsm{} K562 comparison. A seed-$0$ calibration check on
regenerated deep-model predictions did not reproduce the original
predictions to the pre-specified tolerance; we therefore avoid
uncertainty estimates that would mix the original paper predictions
with regenerated deep-model outputs. These intervals are historical uncertainty
evidence for the earlier classical comparisons, not for the locked external
evaluation.

\section{Architecture controls for \texorpdfstring{$\xvec$}{x}-only MLP}
\label{app:no-norm-mlp}

To test whether the $\xvec$-only MLP failure to recover the
magnitude-like signal is explained by normalization, dropout, capacity,
or the absence of a simple scale-preserving path, we ran an
architecture-control audit. We held the data, target, optimizer, and
training schedule fixed and varied only the encoder: (B)~remove
LayerNorm; (C)~remove LayerNorm and Dropout---the backbone already sets dropout to
$p = 0$ (App.~\ref{app:spsm}), so (C) is expected to coincide with (B) and we retain it
only as the audit enumerated it; (D)~widen the MLP to
$1024$ hidden units; (E)~add a learnable scalar shortcut
$f(\xvec) = \mathrm{MLP}(\xvec) + \beta \cdot \|\xvec\|_2$
with $\beta$ initialized at $0$. Variants (B)--(D) are strictly
$\xvec$-only; variant (E) augments $\xvec$ with the single
hand-designed magnitude scalar $\|\xvec\|_2$ (i.e., $m_2$
from Eq.~\ref{eq:mag-features}) routed through an additive shortcut,
so it is the strongest x-with-magnitude-shortcut control we tested.
Across three seeds each, no variant crosses VCC test $\rsq = +0.150$
(best individual seed: wide-no-norm seed $1$ at $\rsq = +0.034$);
means range from $-0.040 \pm 0.021$ to $+0.016 \pm 0.018$ (sample
standard deviation, $\mathrm{ddof}{=}1$). Per a pre-specified rule
(\textsc{Partial-Rescue} if mean $\rsq \geq +0.150$), the outcome is
\textsc{No-Rescue}.
None of these tested variants rescued the MLP under the pre-specified threshold. Removing LayerNorm, widening the network, and the enumerated no-Dropout variant did not rescue the strictly $\xvec$-only models; the additive $m_2$ shortcut also failed to rescue the augmented variant. Because the backbone already has dropout $p=0$, the no-Dropout arm is not an independent test of dropout, and because variant (E) supplies a hand-designed magnitude scalar, it is not an $\xvec$-only control. This audit is
reported as a diagnostic robustness analysis covering four common
architecture variants and is not exhaustive; activation, optimizer,
and skip-connection variants remain untested.

\section{RF hyperparameter robustness}
\label{app:rf-sweep}

To test whether the Random Forest result on $[\xvec;\,\magmaptilde]$
depends on a narrow hyperparameter choice, we ran a targeted
hyperparameter robustness sweep. We held the data, features, and
seeds fixed and varied only the RF: paper reference S0
($n_{\text{est}}{=}200$, $\max_{\text{depth}}{=}20$,
$\min_{\text{leaf}}{=}2$); S1 shallower trees
($\max_{\text{depth}}{=}10$); S2 unbounded depth; S3 stronger leaf
regularization ($\min_{\text{leaf}}{=}5$); S4 more trees
($n_{\text{est}}{=}500$); S5 intermediate depth
($\max_{\text{depth}}{=}15$). Across three seeds $\times$ six
settings ($18$ fits), VCC test $\rsq$ remained in $[+0.361, +0.372]$
(Table~\ref{tab:rf-sweep}), every setting meeting the pre-specified
robustness criterion (VCC $\rsq \geq +0.330$). We report this sweep as a
robustness check and keep the original S0 setting as the paper's primary
RF configuration. (The original sweep also recorded a legacy
distributed-endpoint K562 transfer Spearman per setting; those values
are not reproduced under the locked revision harness and are omitted.)

\begin{table}[h]
\centering
\small
\caption[RF hyperparameter robustness sweep]{RF hyperparameter robustness sweep on $[\xvec;\,\magmaptilde]$.
Mean $\pm$ sample standard deviation of VCC test $\rsq$ across three seeds (ddof=1) for
each setting. All settings meet the robustness criterion used for this sweep (VCC $\rsq \geq +0.330$).}
\label{tab:rf-sweep}
\begin{tabular}{llc}
\toprule
ID & Configuration & VCC test $\rsq$ \\
\midrule
S0 & Paper reference ($n_{\text{est}}{=}200$, depth$=20$, leaf$=2$) & $+0.370 \pm 0.004$ \\
S1 & Shallower trees (depth$=10$) & $+0.361 \pm 0.003$ \\
S2 & Unbounded depth & $+0.369 \pm 0.004$ \\
S3 & Leaf regularization (leaf$=5$) & $+0.367 \pm 0.003$ \\
S4 & More trees ($n_{\text{est}}{=}500$) & $+0.372 \pm 0.003$ \\
S5 & Intermediate depth (depth$=15$) & $+0.370 \pm 0.004$ \\
\bottomrule
\end{tabular}
\end{table}

\section{Additional discussion: \magspsm{} and MAG-Anchor as a secondary deep method}
\label{app:additional-discussion}

\paragraph{Why we exposed magnitude as a feature rather than as a loss.}
Reweighting losses (Balanced MSE \citep{ren2022balancedmse}, LDS
\citep{yang2021delving}) act on the objective while leaving the encoder's input
unchanged. In the exploratory runs of App.~\ref{sec:diagnostic} they did not resolve
prediction-range collapse. Those runs predate the locked evaluation and their
artifacts were not retained, so we treat this as the motivation for supplying
magnitude as an explicit feature rather than as an established mechanism. Reweighting
can in principle reshape a learned representation through its gradients; our claim is
only that the variants we tried did not.

\paragraph{Foundation models scope.}
The diagnostic claims of this section concern the MLP-encoder family and may not
generalize to other foundation-model-based perturbation predictors. Geneformer
is evaluated here as a \emph{frozen} representation
(Sec.~\ref{sec:informativeness}, Sec.~\ref{sec:transfer}); its behavior under
fine-tuning is a separate question that we did not address. Whether other large
pretrained single-cell encoders such as scGPT~\citep{cui2024scgpt} and Universal
Cell Embeddings~\citep{rosen2026uce} recover magnitude on the disjoint
target-gene split, and whether any of these encoders recovers it once its weights
are allowed to adapt, was not evaluated. Compute and preprocessing costs
(gene-vocabulary alignment, and the memory budget for fine-tuning on a single
24\,GB device) placed a fairness-controlled fine-tuning comparison outside the
scope of this paper; it is an important direction for follow-up work.

\paragraph{\magspsm{} as a proof of concept, not the final model.}
The strong-baseline audit (App.~\ref{sec:strong-baselines}) decisively
shows that \magspsm{} is not the strongest exploiter of the magnitude
signal. Random Forest on the same features outperforms it by
$\Delta \rsq = +0.145$ in-distribution. An earlier version of this work also reported a gap on the
legacy distributed endpoint; that value was not re-derived under the locked harness and we
do not use it as evidence here. We retain \magspsm{} in the main text as a
diagnostic deep augmentation: the simplest verification that the same
magnitude signal can be exposed to a deep encoder, and the substrate
for the per-row alignment controls in App.~\ref{sec:causal}. We do not recommend
\magspsm{} as a final architecture.

\paragraph{MAG-Anchor as a secondary deep method.}
MAG-Anchor was not reproduced under the locked revision harness. We retain its
architecture and earlier controls (App.~\ref{app:mag-anchor-full}) for provenance
only and make no current quantitative or transfer claim from them. Random Forest
on $[\xvec;\,\magmaptilde]$ remains the stronger reproduced in-distribution
model; no locked-evaluation conclusion depends on MAG-Anchor.

\paragraph{Two readings of the diagnostic.}
A reasonable alternative reading of our finding is that mean-collapse
arises from optimization or normalization behavior of the deep MLP when
the label scale couples to input scale through normalization layers,
rather than from a representational property of perturbation prediction
\emph{per se}. The architecture-control analysis in
App.~\ref{app:no-norm-mlp} tests this reading directly: holding the
data, target, optimizer, and training schedule fixed and varying only
the encoder (no LayerNorm; no LayerNorm and no Dropout; widened to
$1024$ hidden units; an additive scale-preserving shortcut), no
$\xvec$-only variant crosses VCC test $\rsq = +0.150$, well below the
$+0.208$ obtained by a four-feature linear regression on
$\widetilde{\boldsymbol{m}}(\xvec)$. This empirically disfavors the
optimization-only reading within the MLP family we tested, though it
does not exclude it for architectures we did not test (set
transformers, attention-based encoders, foundation-model fine-tuning).
Within the MLP-family diagnostic, explicit magnitude exposure separates the tested models that recover more of the low-dimensional predictive signal from those that collapse. We do not turn that observation into a general feature-engineering recommendation: Sec.~\ref{sec:endpoint-depth} shows that the VCC aggregate-magnitude signal is substantially entangled with sampling depth, and Sec.~\ref{sec:magnitude} shows that adding the same block does not rescue zero-shot transfer under the frozen primary head. The diagnostic is therefore most precisely stated as a finding about \emph{which tested models, on the disjoint target-gene split with this label parameterization, exploit a low-dimensional predictive signal whose biological interpretation is not cleanly separable from benchmark design}. 

\section{Additional limitations}
\label{app:additional-limitations}

\begin{enumerate}[itemsep=2pt,topsep=2pt,leftmargin=*]
\item \emph{The label is magnitude-like by construction.} Because the
target is the log Anderson--Darling distance from NT, magnitude
features are in principle informative. Our claim is not that this
relationship is biologically surprising in isolation; it is that
direction is fragile under the disjoint target-gene split, that
the tested MLP encoders fail to exploit the aggregate magnitude block reliably in-domain, and that explicit exposure changes performance in those diagnostics. This is not evidence that the block is a clean biological response variable: Sec.~\ref{sec:endpoint-depth} shows substantial overlap with sampling depth on VCC, and Sec.~\ref{sec:magnitude} shows that the block does not rescue the frozen primary zero-shot transfer result.
\item \emph{Replogle preprocessing residuals.} Replogle K562 uses a
different per-row biological unit than VCC ($1{,}430$ of $2{,}000$
HVGs mapped; feature scales differ by $\sim 1.9\times$). We align via
VCC-side normalization (App.~\ref{app:replogle-mag}); residual
confounding cannot be fully ruled out.
\item \emph{Cross-dataset $\rsq$ scale disparity.} VCC test, Replogle
K562 transfer, and Replogle RPE1 transfer have AD-based labels with
substantially different marginal standard deviations (sample std,
$\mathrm{ddof}{=}1$): VCC $\sigma_y = 0.758$, K562 $\sigma_y = 2.750$
($3.63\times$ the VCC value), and RPE1 $\sigma_y = 2.372$
($3.13\times$ the VCC value). Since $\rsq$ denominators scale with
the squared label std, cross-dataset $\rsq$ magnitudes are not
directly comparable across the three benchmarks; the K562 $\rsq$
denominator is approximately $13.2\times$ the VCC denominator and the
RPE1 denominator is approximately $9.8\times$. Cross-dataset $\rsq$
values should therefore be read as within-dataset goodness-of-fit
measures, not as a single calibrated quantity transferable from one
dataset to the other; rank-based metrics (Spearman $\rho$) are
unaffected by this scale disparity and are the metric we emphasize in
cross-cell claims.
\item \emph{HVG-coverage confound on cross-cell transfer.} Models are
trained on the full $2{,}000$-dimensional VCC HVG namespace. In the
primary Replogle K562 transfer evaluation, Replogle rows are embedded
into this $2{,}000$-dimensional namespace by zero-imputing the $570$
VCC-only coordinates missing from Replogle. This is a structural input
shift that is logically distinct from the biological cross-cell shift.
The HVG-coverage control in App.~\ref{app:hvg-control} removes the
zero-padding by training and evaluating directly in the $1{,}430$
shared-HVG subspace, and K562 rank association is preserved and slightly
higher ($\rho = +0.377 \pm 0.010$ vs.\ $+0.314 \pm 0.009$). This
addresses the coverage question only for the legacy distributed-endpoint
analysis; the corresponding prediction vectors were not retained, so the
control cannot be re-scored against the rebuilt target-gene endpoint
without an unvalidated refit.
\item \emph{Modest transfer magnitudes.} On the distributed Replogle
endpoint the fits we observe are modest (legacy distributed-endpoint
values are not reproduced under the locked revision harness and are
omitted). On the rebuilt target-gene endpoint we report rank
association rather than pooled $\rsq$, since predictions produced on one
outcome scale are not calibrated to another. The magnitude increment is
modest and does not solve cross-context generalization.
\item \emph{Tree behavior on low-dimensional input.} On VCC, Random Forest on
\textsc{mag-only} has test $\rsq=-0.042$ but Spearman $+0.555$, indicating
in-domain rank information with poor calibration. We do not use the unreproduced
legacy Replogle tree values to support a current transfer claim.
\item \emph{No exhaustive deep-architecture search.} The trained encoders we
tested are MLP-family (\spsm{}/\magspsm{}, MAG-Anchor, and the MAGAnchor-x
ablation); Geneformer enters only as a frozen representation, with no encoder
parameter updated. Set Transformers, attention-based encoders, foundation-model
\emph{fine-tuning} (scGPT~\citep{cui2024scgpt},
Geneformer~\citep{theodoris2023geneformer}, UCE~\citep{rosen2026uce}), and
width and depth sweeps were not evaluated. The conclusion that no deep encoder we
trained matches the tree-ensemble baseline is therefore empirical and limited to
the architectures listed. Within the MLP family, additional architecture controls
(no-LayerNorm, no-Dropout, wider, and scale-aware variants) did not
recover signal from $\xvec$-only input
(App.~\ref{app:no-norm-mlp}). Whether other
architecture families recover magnitude from $\xvec$ alone, or
whether the gap to RF on $[\xvec;\,\magmaptilde]$ narrows under such
architectures, remains an important open question.
\item \emph{MAG-Anchor is historical context only.} It was not reproduced under
the locked revision harness, so its earlier numerical gaps and external
associations are not used as evidence. Its architecture and controls are retained
for provenance; no locked-evaluation conclusion depends on it.
\item \emph{Statistical reporting choices and their scope.} We report
all stochastic results as mean $\pm$ sample standard deviation
($\mathrm{ddof}{=}1$) across seeds, and report paired three-seed
deltas for matched-architecture comparisons (e.g.,
\magrank{}~vs.~\magspsm{}, $\Delta\rsq = +0.0142 \pm 0.0040$;
App.~\ref{app:rank}). App.~\ref{app:bootstrap-ci} separately reports three
row-level bootstrap intervals from the earlier classical analysis
($n_{\text{VCC}}{=}3{,}972$, $n_{\text{K562}}{=}2{,}176$); these do not govern
the locked external evaluation. We do not report formal hypothesis tests,
paired Wilcoxon signed-rank statistics, or multiple-comparison
corrections across the full method grid; with
$n_{\text{seeds}}{\in}\{3,5\}$, such tests would be underpowered for
small effects, and we judged that qualitative pattern descriptions
are more honest than artificially sharp $p$-values. The central qualitative findings of the paper do not depend on
small-effect comparisons: H-FM1--H-FM3 use their pre-specified bootstrap
interval criteria rather than a $p$-value cutoff, while the sampling-depth primary
is explicitly descriptive and non-causal. The legacy intervals above are not
needed for either locked primary. Pairwise
deltas with overlapping intervals are flagged as unresolved in the
Discussion and
in App.~\ref{app:additional-discussion}; we do not claim these are
statistically distinguishable. A targeted hyperparameter robustness
sweep for the RF result is reported in App.~\ref{app:rf-sweep}.
\end{enumerate}

\section{Anderson--Darling label construction}
\label{app:ad-label}

For each row $i = (b_i, g_i)$ with experimental Flex batch $b_i$ and
target gene $g_i$, the regression label is
\[
y_i \;=\; \log\!\bigl(1 + \max\{0,\, A^2_{b_i, g_i}\}\bigr),
\]
where $A^2_{b, g}$ is the two-sample Anderson--Darling statistic
returned by \texttt{scipy.stats.anderson\_ksamp} \citep{scholz1987k}
applied to two one-dimensional samples: (i)~the log1p-normalized
expression values of the target gene $g$ in cells perturbed at $g$
within batch $b$, and (ii)~the log1p-normalized expression values of
the same gene $g$ in non-targeting (NT) control cells from the same
batch $b$. The Anderson--Darling test is therefore applied to the
\emph{target-gene column only}; the $2{,}000$-HVG vector $\xvec_i$ is
the predictor input and does not enter label construction. Negative
AD statistics (rare, possible with the AD null-bias subtraction) are
clamped to zero before the $\log(1+\cdot)$ transform. This per-row
$(b,g)$-indexed convention, with $b$ playing the role of the Replogle
gemgroup-style batch identifier, is the one used in the earlier
distributed-endpoint analysis. The harmonized endpoint that the locked transfer
evaluation of Sec.~\ref{sec:transfer} scores against is built differently: it is
construct-level, uses a pooled non-targeting control, and is computed by repeated
fixed-size subsampling (App.~\ref{app:harmonized-endpoint}).

\section{Results from an earlier version of this work, not re-derived}
\label{app:legacy-not-rederived}
\label{app:harmonized-transfer-full}
The tables in this section reproduce results from the earlier version of this work, computed
with classical baselines. Their numeric values were not re-derived under the locked evaluation
described in Sec.~\ref{sec:protocol}, and the artifacts required to reproduce them were not
retained. They are
included for continuity with the prior version. No claim in this paper depends on them. Each
table's caption states the endpoint against which it was computed.

\begin{table}[t]
\centering
\caption[Legacy classical-model transfer results]{Legacy classical-model transfer results, retained for
historical comparison. These values are not the current locked foundation-model evaluation
(Table~\ref{tab:fm_external_primary}) and were not independently re-derived in the revision harness.
Zero-shot transfer to two external screens on a common
target-gene endpoint. Entries are construct-level Spearman correlations
between each VCC-trained prediction and the harmonized endpoint.
Verdicts follow a pre-specified rule: \textsf{Pos} if the $95\%$
target-gene cluster-bootstrap interval lies wholly above zero,
\textsf{Neg} if wholly below, \textsf{Inc} if it includes zero; an
inconclusive verdict is not evidence of absent transfer. Panel~A
families are evaluated per seed and averaged ($3$ seeds for HistGB and
RF, $1$ for Ridge). Panel~B predictors have a single training seed, so
their intervals are conditional on it. Feature sets: $\xvec$ expression
only, $\magmaptilde$ magnitude only, $[\xvec;\magmaptilde]$ both.
Interval widths are not comparable across screens: the two evaluations
differ in cluster structure (App.~\ref{app:harmonized-transfer-full}).}
\label{tab:harmonized_transfer}
\small
\setlength{\tabcolsep}{5pt}
\begin{tabular}{llcccc}
\toprule
& & \multicolumn{2}{c}{RPE1} & \multicolumn{2}{c}{K562-essential} \\
\cmidrule(lr){3-4}\cmidrule(lr){5-6}
Model & Features & $\rho$ & Verdict & $\rho$ & Verdict \\
\midrule
\multicolumn{6}{l}{\emph{Panel A: classical families, seed-averaged}} \\
Ridge   & $\xvec$                 & $-0.091$ & Neg & $+0.009$ & Inc \\
Ridge   & $\magmaptilde$          & $+0.097$ & Pos & $+0.133$ & Pos \\
Ridge   & $[\xvec;\magmaptilde]$  & $-0.068$ & Neg & $-0.043$ & Inc \\
HistGB  & $\xvec$                 & $-0.116$ & Neg & $-0.211$ & Neg \\
HistGB  & $\magmaptilde$          & $+0.009$ & Inc & $+0.160$ & Pos \\
HistGB  & $[\xvec;\magmaptilde]$  & $-0.038$ & Inc & $+0.018$ & Inc \\
RF      & $\xvec$                 & $-0.105$ & Neg & $-0.218$ & Neg \\
RF      & $\magmaptilde$          & $+0.053$ & Pos & $+0.178$ & Pos \\
RF      & $[\xvec;\magmaptilde]$  & $-0.018$ & Inc & $+0.166$ & Pos \\
\midrule
\multicolumn{6}{l}{\emph{Panel B: deep and magnitude-only predictors}} \\
\spsm{} baseline & ---             & $-0.050$ & Neg & $-0.070$ & Neg \\
\magspsm{}       & ---             & $+0.067$ & Pos & $+0.016$ & Inc \\
\maglinear{}     & $\magmaptilde$  & $+0.057$ & Pos & $+0.147$ & Pos \\
\bottomrule
\end{tabular}
\end{table}

Table~\ref{tab:harmonized_transfer} reports point estimates and
verdicts. The corresponding $95\%$ intervals come from a paired cluster
bootstrap over target genes with $B = 10{,}000$ replicates and a
pre-specified seed. Each replicate samples target-gene clusters with
replacement; a cluster drawn more than once contributes all of its
constructs once per occurrence; and the same draw is applied jointly to
every prediction, every seed, and every contrast within that replicate,
so that paired comparisons share resampling noise.

For the nine classical families the metric is computed per seed and
averaged, with the averaging repeated inside each replicate; for the
three single-seed predictors the metric comes from the one archived
prediction vector, so the interval reflects target-gene sampling
conditional on that seed. No replicate was discarded for an undefined
correlation in either screen.

The two evaluations differ in cluster structure: RPE1 has $1{,}932$
constructs over $1{,}811$ target genes with at most three constructs per
gene, and K562-essential has $1{,}903$ over $1{,}811$ with at most two. These are distinct
screen-specific gene sets of equal cardinality.
Interval widths are therefore not mechanically comparable across
screens.

\subsection{HVG-coverage control}
\label{app:hvg-control}

To test whether the legacy K562 distributed-endpoint transfer result
is explained by the
$2{,}000$-to-$1{,}430$ HVG dimension drop, we reran the RF
$[\xvec;\,\magmaptilde]$ baseline using only the $1{,}430$ HVGs shared
between VCC and Replogle K562. The shared genes were selected
deterministically by gene name in VCC HVG order, without using
Replogle target labels or outcomes. Magnitude features were
recomputed from the $1{,}430$-dimensional inputs and $z$-scored using
VCC training statistics only. Across three seeds with the same RF
hyperparameters as the main experiment ($n_{\text{est}}{=}200$,
$\max_{\text{depth}}{=}20$, $\min_{\text{leaf}}{=}2$), the
$1{,}430$-HVG control achieved VCC test $\rsq = +0.340 \pm 0.002$ and
K562 transfer Spearman $\rho = +0.377 \pm 0.010$
(Table~\ref{tab:hvg_control}). This passes our pre-specified
robustness rule ($\rho \geq +0.20$) and shows that, \emph{when transfer
is scored against the distributed endpoint}, the positive K562 rank
association is not explained by the missing $570$ HVGs
alone. The in-domain VCC $\rsq$ decreases relative to the
$2{,}000$-HVG result ($+0.370 \pm 0.004$), as expected from the
reduced feature set, but cross-cell rank transfer is preserved.

\begin{table}[t]
\centering
\caption[HVG-coverage control for K562 transfer]{HVG-coverage control for K562 transfer against the
legacy distributed-breadth endpoint. Mean $\pm$ sample standard deviation over three seeds.}
\label{tab:hvg_control}
\small
\begin{tabular}{lcc}
\toprule
Training namespace & VCC test $\rsq$ & K562 Spearman $\rho$ \\
\midrule
$2{,}000$ VCC HVGs & $+0.370 \pm 0.004$ & $+0.314 \pm 0.009$ \\
$1{,}430$ shared HVGs & $+0.340 \pm 0.002$ & $+0.377 \pm 0.010$ \\
\bottomrule
\end{tabular}
\end{table}

\paragraph{Pre-specified HVG-coverage control: $\xvec$-only models.}
We additionally pre-specified a symmetric x-only counterpart of the
control above to test whether the negative cross-cell transfer of
expression-only models is a feature-truncation artifact rather than a
direction-vs-magnitude phenomenon. The pre-registration committed four
hypotheses with PASS/WARN/FAIL thresholds and the consequent paper-text
edits before any model was trained
in the pre-registration file committed to the project repository.
We trained Ridge, Random Forest, HistGradientBoosting, and the deep
\spsm{} V4 baseline on $\xvec$ alone restricted to the same $1{,}430$
shared HVGs, three seeds each, with hyperparameters frozen at their
main-paper values. Three-seed mean K562 Spearman values were:
Ridge $\rho = -0.229$, Random Forest $\rho = -0.407 \pm 0.003$,
HistGradientBoosting $\rho = -0.305 \pm 0.019$, and \spsm{} V4 baseline
$\rho = -0.389 \pm 0.007$. All four pre-specified hypotheses passed:
H1 (every $\xvec$-only classical model retains negative mean Spearman)
passed with the largest value at Ridge $-0.229$;
H2 ($\xvec$-only deep baseline mean $\rho < +0.05$) passed at $-0.389$;
H3 (best magnitude-aware minus best $\xvec$-only $\rho$ gap on the
$1{,}430$-HVG sub-space $\geq +0.20$) passed with gap $+0.607$
($+0.377$ vs $-0.229$); and H4 (within-VCC $|\Delta\rsq| \leq 0.030$
relative to the $2{,}000$-HVG baselines) passed with worst-case
$|\Delta\rsq| = 0.011$ for Ridge. Per the pre-specified decision rule,
the legacy distributed-endpoint $\xvec$-only negative-transfer pattern
is robust to
feature truncation and the magnitude--direction asymmetry holds on the
shared-HVG sub-space.

\paragraph{Scope.}
Per-construct prediction vectors from this control were not retained,
so it cannot be re-scored against the rebuilt target-gene endpoint
without an unvalidated refit. We therefore treat it as a legacy
distributed-endpoint sensitivity analysis and do not use it to support
Table~\ref{tab:harmonized_transfer}.

\end{document}